\documentclass[twoside,11pt]{article}

\usepackage[preprint]{jmlr2e} 

\usepackage{url}
\usepackage{amsmath}
\usepackage{booktabs}
\usepackage[accsupp]{axessibility} % Improves PDF readability
\usepackage{adjustbox}
\usepackage{multirow} 
\usepackage{ragged2e}
\usepackage{marvosym}
\usepackage{pifont}
\usepackage{tabularx}
\usepackage{wrapfig}
\usepackage{caption}
\usepackage{subcaption}

\usepackage[switch]{lineno}

\usepackage[table]{xcolor}
\usepackage{colortbl}

\definecolor{darkred}{rgb}{0.6, 0, 0}
\definecolor{darkgreen}{rgb}{0, 0.6, 0}
\definecolor{lightblue1}{RGB}{222,235,247} % lightest blue
\definecolor{lightblue2}{RGB}{189,215,238} % medium blue
\definecolor{lightblue3}{RGB}{158,202,225} % slightly darker blue

\newcommand{\cmark}{\textcolor{darkgreen}{\ding{51}}} % Checkmark
\newcommand{\xmark}{\textcolor{darkred}{\ding{55}}} % Cross
\newcommand{\na}{\text{---}} % For N/A or unmentioned data

\usepackage{lastpage}
\jmlrheading{25}{2025}{1-\pageref{LastPage}}{1/25; Revised X/25}{X/25}{25-0000}{Nisarg,Vinayak and Li-Yun}

\ShortHeadings{MODEST: Multi-Optics Depth-of-Field Stereo Dataset}{Nisarg,Vinayak and Li-Yun}
\firstpageno{1}

\begin{document}

% Uncomment the following line if you want line numbers for review
% \linenumbers 

\title{MODEST: Multi-Optics Depth-of-Field Stereo Dataset}

\author{\name Nisarg K. Trivedi* \email nisarg819@gmail.com \\
       %\addr California, USA
       \AND
       \name Vinayaka A. Belludi \email belludivinayakaa@gmail.com \\
       %\addr Karnataka, India 
       \AND
       \name Li-Yun Wang \email wangliyun0117@gmail.com \\
       %\addr California, USA
       }

\editor{Pending Editor}

\maketitle

\begin{abstract}%   <- trailing '%' for backward compatibility of .sty file
% original full length abstract preserved for Arxiv
Training and evaluation of state-of-the-art computer vision algorithms for reliable shallow depth of field (DoF) rendering and defocus deblurring remain constrained by a persistent lack of large-scale, full-frame, high fidelity, real-image datasets. Optical effects of shallow depth of field and defocus blur depend intimately on camera optical configuration set with focal length and aperture; requiring rigorous evaluation of the models when these parameters systematically change. Further, modern applications such as Augmented, Virtual Reality (AR, VR), smartphones, industrial robots, etc. deploy stereo or multi-camera systems. We present MODEST - the first ultra-high-resolution (5472×3648px, 20MP), multi-optics depth of field stereo DSLR dataset that methodically varies focal length and aperture for a series of complex, real-world scenes, capturing the optical realism and complexity of professional camera systems. With 20,000 images across 50 distinct optical configurations, focal length in 28-70mm and aperture in f/22-f/2.8 for multiple stereo viewpoints for 10 scenes, this ultra-high-resolution, full-range optics coverage enables controlled analysis of geometric and optical effects for shallow depth of field rendering and defocus deblurring algorithms. Each scene is intentionally curated to include challenging visual elements, including reflective surfaces, transparent glass walls, fine-grained details, point lights, and multi-scale depth illusions. In addition, we provide image sets for monocular intrinsics calibration and stereo extrinsics calibration for each focal configuration to support ever-evolving classical and learning-based calibration methods. We evaluate several state-of-the-art algorithms for depth of field and defocus deblurring across focal configurations and demonstrate failure cases and limitations. With comprehensive tuning analysis, we demonstrate how MODEST can evaluate sensitivity of the SOTA DoF models for the actual focal configuration of input images. This work attempts to bridge the gap between synthetic, low-resolution training data and inference generalization on high-resolution, real camera optics. We release the dataset, calibration files, and evaluation code to support reproducible benchmarking and further research on real-world optical generalization.\footnote{* Correspondence author. Dataset and software tools are available for non-commercial and purely academic research use at \url{https://modest-dataset.netlify.app/}}

\end{abstract}


% ---------------------------------------------------------------
% Main Body Sections
% \IEEEraisesectionheading{
    \section{Introduction}\label{sec:intro}
% }

% add defocus and deblurring datasets.
% number of images should be checked.

% \textcolor{green}{
% datasets in dof can be used in deblur methods as well...?
% }
\begin{figure}[htbp]  % * makes it span both columns
    \centering
    \setlength{\abovecaptionskip}{2pt}  % reduce space above caption
    \setlength{\belowcaptionskip}{2pt}  % optional: reduce space below caption
    \includegraphics[width=\textwidth]{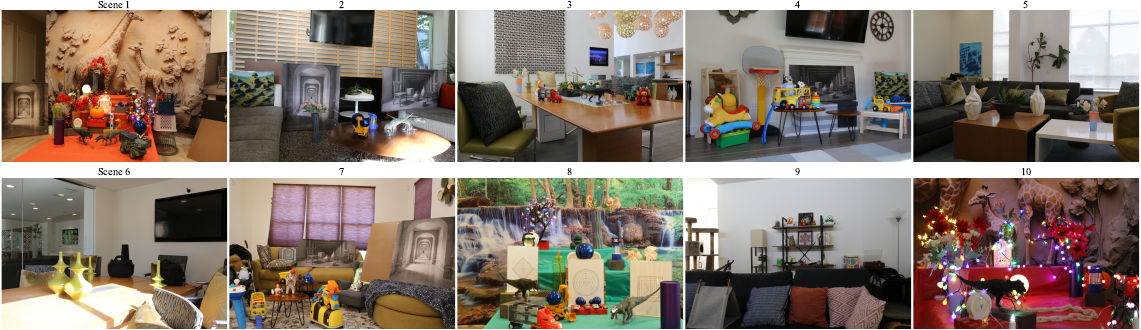}  \\
    % scale to text width
    \vspace{-2pt} 
    \caption{MODEST dataset. For 10 scenes with varying scene complexity, lighting and background, images are captured with two identical camera assemblies at 10 focal lengths (28-70mm) and 5 apertures f/2.8 - f/22.0 in 2000 images per scene. The data features challenging elements such as multi-scale optical illusions, reflective surfaces, transparent glass doors, sharp lighting changes, ambiguous background depths. Besides a global calibration set, each scene and each focal length has dedicated calibration sets enabling use of classical and learning-based calibration methods for intrinsics and extrinsics.}
    \label{fig:data}
    \vspace{-2pt}
\end{figure}

Shallow depth of field (DoF) rendering is the process of mimicking the optical blur in a wide aperture camera. Shallow DoF guides our view towards a narrow sharp region of the image with other regions blurred based on their distance from the focus plane. With synthetically mimicked optical blur, researchers from Google develop the effects of shallow DoF on their smartphone~\citep{wadhwa2018synthetic}. It allows users to render DoF images efficiently and automatically on smart devices without professional photography knowledge. Additionally, shallow DoF rendering is a technique for helping users focus on sharp virtual objects in AR applications~\citep{kim2023extended, shi2021augmented}.

Unlike synthetic blur, real optical defocus is obtained by a complex interaction between aperture, focal length, sensor size, focus distance, lens characteristics, and scene geometry. Consequently, images captured with different focal lengths and aperture settings exhibit substantially different blur distributions, bokeh characteristics, and depth transitions, making accurate shallow DoF rendering and defocus deblurring considerably more challenging via machine learning methods. While recent learning-based methods have demonstrated impressive performance, their ability to generalize across diverse optical configurations is largely constrained by the limited variations in the existing datasets. Most public datasets provide only a single focal length or a limited set of camera settings, preventing models from learning the rich optical behavior encountered in real DSLR imaging systems.

To address these limitations, a dataset can be captured under a systematic acquisition protocol that explicitly varies camera optics while maintaining consistent scene geometry. Such controlled image acquisition enables fair evaluation of algorithms across different focal lengths and aperture settings and facilitates the study of how optical parameters influence rendering and restoration performance. Furthermore, comprehensive camera calibration images can be captured for each focal configuration, which is essential for computational photography tasks including stereo matching and future camera calibration research. Therefore, alongside the captured images, calibration sequences should be provided to enable researchers to recompute intrinsic and extrinsic camera parameters using future calibration algorithms without requiring the original capture setup.

\begin{table}
\centering
\caption{Data comparison categorized for different use cases, including shallow depth of field rendering and defocus deblurring and deblurring. (\dag, \ddag): systematic focal length and aperture variation. RGB-P: RGB with a polarisation sensor. Active Stereo: an active stereo depth sensor. \vspace{-8pt}} %D-ToF, I-TOF: direct, indirect Time-of-Flight sensor.
\label{table:dataset_comparison}
\footnotesize % Use a smaller font size for a wide table

\begin{adjustbox}{width=\textwidth}
\begin{tabular}{l c c c c c c c c c}
    \toprule
    \textbf{Dataset} & \textbf{Capture} & \textbf{Real/} & \textbf{Scene} & \textbf{Number of} & \textbf{Resolution} & \textbf{Focal Len.} & \textbf{Aperture} & \textbf{Depth} & \textbf{Calibration} \\
    & \textbf{Setup} & \textbf{Synthetic} & \textbf{Number} & \textbf{Images} & {px} & \textbf{Variation \dag} & \textbf{Variation \ddag} & \textbf{Range} & \textbf{Set} \\
    \midrule
    \multicolumn{10}{l}{\textbf{Use Case: Shallow dof rendering and defocus deblurring}} \\
    \cmidrule{1-10}
    CUHK ('14)~\citep{Shi2014BlurDetection} & Internet Images + Annotated Binary Blur Masks & Real & \na & 1000 & $352 \!\times\! 352$ & \na & \na & \na & \xmark \\
    RTF ('16)~\citep{d2016non} & Mono RGB (Lytro Light‑field Camera) & Real & 22 & 44 & $360 \!\times\! 360$ & \xmark & \cmark & \na & \xmark \\
    DPDD ('20)~\citep{abuolaim2020defocus} & Mono RGB & Real & 500 & 2000 & $6720 \!\times\! 4480$ & \xmark & \xmark & $0.3\text{m} - 10\text{m}$ & \xmark \\
    LFDOF ('21)~\citep{ruan2021aifnet} & Mono RGB (Light-field Camera) + LiDAR & Real & \na & 23978 & $375 \!\times\! 540$ & \xmark & \xmark & $0.3\text{m} - 10\text{m}$ & \xmark \\
    RealDOF ('21)~\citep{lee2021iterative} & Stereo RGB (Dual-camera with a beam splitter) & Real & 50 & 100 & $1536 \!\times\! 2320$ & \xmark & \cmark & \na & \xmark \\
    BLB ('22)~\citep{bokehme_blb} & Synthetic renderings from Blender & Synth & 10 & 1000 & $1920 \!\times\! 1080$ & \xmark & \cmark & $0.5\text{m} - 10\text{m}$ & \xmark \\
    VABD ('24)~\citep{chen2024variable} & Mono RGB & Real & 535 & 2940 & $1536 \!\times\! 1024$ & \xmark & \cmark & \na & \xmark \\
    RealBokeh ('25)~\citep{seizinger2025bokehlicious} & Mono RGB (Canon EOS R6 II) & Real & 300 & 23000 & $6000 \!\times\! 4000$ & \cmark & \cmark & \na & \xmark \\
    MODEST (Ours) & Stereo RGB (Canon 6D) & Real & 9 & 20000 & $\mathbf{5472 \!\times\! 3648}$ & \cmark & \cmark & $0.5\text{m} - 10\text{m}$ & \cmark \\
    \midrule
    \multicolumn{10}{l}{\textbf{Use Case: Deblurring}} \\
    \cmidrule{1-10}
    %GoPro ('17)~\citep{nah2017deep} & Mono RGB (GoPro Hero4 Black Camera) & Synth & 33 & 3214 & $1280 \!\times\! 720$px & \xmark & \xmark & \na & \xmark \\
    DeepLens ('18)~\citep{lijun2018deeplens} & Mono RGB (Dual-lens Camera + Thin-lens model and Billboards) & Real + Synth & \na & 502462 & $3024 \!\times\! 4032$ & \xmark & \cmark & \na & \xmark \\
    %HIDE ('19)~\citep{shen2019human} & Mono RGB (Gopro Hero4 Black Camera) & Synth & 31 & 16844 & $1280 \!\times\! 720$px & \xmark & \xmark & $0\text{m} - 3\text{m}$ & \xmark \\
    %REDS ('19)~\citep{nah2019ntire} & Mono RGB (GoPro Hero6 Black Camera) & Synth & 300 & 30000 & $1280 \!\times\! 720$px & \xmark & \xmark & $0\text{m} - 3\text{m}$ & \xmark \\
    SYNDOF ('19)~\citep{lee2019deep} & Mono RGB (Thin-lens model) & Synth & \na & 8231 & (960-2000) $\!\times\!$ (436-2000) & \cmark & \cmark & $0.5\text{m} - 80\text{m}$ & \xmark \\
    %RealBlur ('20)~\citep{rim2020real} & Mono RGB (Dual-camera System with a Beam Splitter) & Real & 232 & 18952 & $680 \!\times\! 772$px & \xmark & \xmark & \na & \cmark \\
    BSD-B ('20)~\citep{zhong2020efficient} & Mono RGB (Convolving Sharp Images with Uniform Disk Kernels) & Synth & 500 & 40000 & (320-480) $\!\times\!$ (320-480) & \na & \cmark & \na & \na \\
    %NTIRE ('21)~\citep{nah2021ntire} & Mono RGB (GoPro Hero6 Black Camera) & Synth & 300 & 30000 & $1280 \!\times\! 720$px & \xmark & \xmark & $0.5\text{m} - infinity$ & \xmark \\
    %RSBlur ('22)~\citep{rim2022realistic} & Mono RGB (Dual-camera system) & Real & 697 & 13358 & $1920 \!\times\! 1200$px & \xmark & \xmark & $1\text{m} - infinity$ & \xmark \\
    iDFD ('23)~\citep{nazir2023idfd} & Mono RGB (Dual-sensor Rig) + Microsoft Kinect (Depth) & Real & \na & 1528 & $1050 \!\times\! 1050$ & \xmark & \cmark & $0\text{m} - 10\text{m}$ & \xmark \\
    %GS-Blur ('24)~\citep{lee2024gs} & Mono RGB (3D Gaussian Splatting Model) & Synth & 18542 & 312418 & $1280 \!\times\! 720$px & \cmark & \xmark & \na & \xmark \\
    QPDD ('25)~\citep{chen2025quad} & Mono RGB (50-megapixel Quad-Pixel Sensor) & Real & 300 & 9870 & (1080-4096) $\!\times\!$ (1280-3072) & \xmark & \cmark & $0.2\text{m} - \!\infty\!$ & \xmark \\
    MODEST (Ours) & Stereo RGB (Canon 6D) & Real & 10 & 20000 & $\mathbf{5472 \!\times\! 3648}$ & \cmark & \cmark & $0.5\text{m} - 10\text{m}$ & \cmark \\
    %\midrule
    %\multicolumn{10}{l}{\textbf{Use Case: Optical illusions}} \\
    %\cmidrule{1-10}
    %3D-Visual-Illusion ('25)~\citep{yao20253d} & Stereo RGB + LiDAR & Real + Synth & 3000 & \na & $1080 \!\times\! 1920$px & \na & \na & $0.5\text{m}-50\text{m}$ & \xmark \\
    %MonoTrap ('25)~\citep{bartolomei2025stereo} & Stereo Image RGB + LiDAR & Real & 26 & \na & $585 \!\times\! 375$px & \xmark & \xmark & $0\text{m} - 10\text{m}$ & \xmark \\
    %MODEST (Ours) & Stereo RGB (Canon 6D) & Real & 9 & 20000 & $\mathbf{5472 \!\times\! 3648}$px & \cmark & \cmark & $0.5\text{m} - 10\text{m}$ & \cmark \\
    \bottomrule
\end{tabular}
\end{adjustbox}
\vspace{-6pt}
\end{table}

\begin{table}[htbp]
\centering
\caption{MODEST complements recent bokeh datasets.\vspace{-8pt}}
\label{tab:multi_capture}
\resizebox{\columnwidth}{!}{%
\begin{tabular}{l c c c c c}
\toprule
& \textbf{MODEST (ours)} & RealBokeh~\citep{seizinger2025bokehlicious} & EBB!~\citep{ignatov2020rendering} & Aperture~\citep{zhang2019synthetic} & BEDT~\citep{conde2023lens} \\
\midrule
\# Samples & \textbf{20,000} & 23,000 & 4694 & 2942 & 20,000 \\
% \# Train & \na & 20,500 & 4400 & 2942 & 20,000 \\
% \# Validation & \na & 1,250 & 294 & \na & \na \\
% \# Test & \na & 1,250 & \na & \na & \na \\ %N: irrelevant until we do training, removing.
Apertures & \textit{\textbf{f/22.0 - f/2.8}} & \textit{f/20.0 - f/2.0} & \textit{f/1.8} & \textit{f/8.0, f/2.0} & \textit{f/2.0, f/1.8} \\
Focal length & \textbf{28mm - 70mm} & 28mm - 70mm & 85mm & - & - \\
Mono / stereo & \text{Stereo} & Mono & Mono & Mono & Mono \\
Systematic focal length capture & \cmark & \xmark & \xmark & - & - \\ %N: realbokeh should be xmark, isn't it, do they do it systematically? -> L:They did not mention it in their paper. I have checked the paper. N:so it should be a cross, not check mark. They do not do systematic fl.
Systematic aperture capture & \cmark & \xmark & \xmark & \xmark & \xmark \\
Multi-capture & \cmark & \xmark & \xmark & \xmark & \xmark \\
Intrinsic calibration images & \cmark & \xmark & \xmark & \xmark & \xmark \\
Extrinsic calibration images & \cmark & \xmark & \xmark & \xmark & \xmark \\
\bottomrule
\end{tabular}
}
\vspace{-8pt}
\end{table}

In contrast to DoF, defocus deblurring is a process that utilizes image sharpening algorithms to restore high-frequency information lost in the blurred regions caused by a lens being out of focus. Due to this, defocus deblurring is compelling to computer vision applications, including autonomous driving and robotics~\citep{shah2025impact}, surveillance and security~\citep{hillaire2008using}. For autonomous driving and robotics, deblurring algorithms ensure sharp regions for safe navigation. Defocus deblurring algorithms are also used in forensic analysis to reconstruct sharpened regions from blurred images.

Reliable shallow DoF rendering and defocus deblurring under real optical conditions remain challenging for state-of-the-art DoF and defocus deblurring models due to the lack of large-scale, full-frame resolution, high-fidelity, real DSLR datasets. Table \ref{table:dataset_comparison} shows dataset comparisons among different datasets. Most early datasets contain large numbers of scenes from indoor and outdoor environments; however, these datasets encounter several limitations, including low-resolution images and the lack of systematic variations of focal length and apertures. DPDD~\citep{abuolaim2020defocus} is a recent 6K-resolution dataset for model training and evaluation in defocus deblurring, but this dataset does not systematically cover a varying range of focal lengths and apertures. QPDD~\citep{chen2025quad} is another defocus deblurring dataset captured by a camera with sub-aperture views, allowing models to use these sub-aperture views to deblur a pixel. The limitation of the sub-aperture views is the micro-parallax problem, where it is difficult to distinguish between actual depth and sensor noise due to the tiny physical distance between the sub-pixels. Additionally, the existing datasets do not provide calibration sets for recalculating calibration parameters through newer algorithms. Thus, we provide a global intrinsic calibration set and per-focal-length extrinsic calibration sets, allowing recalibration flexibility.

\subsection{Contributions}\label{subsec:contribution}

1) We introduce the first ultra-high resolution (20MP), multi-optics stereo DSLR dataset MODEST containing 20,000 images across 10 scenes; each scene captured with 10 focal lengths and 5 apertures per focal length at multiple stereo viewpoints. Scenes are carefully curated to include challenging visual elements such as optical illusions, reflective and semi-transparent surfaces, ambiguous background depths, point lights, etc. (Fig. ~\ref{fig:data_comps}a). MODEST also contains a global calibration set for monocular intrinsics and a per-scene calibration set for stereo extrinsics, to enable adaptation with ever-evolving calibration methods.
  
\hspace{-1.5em}2) We present the first ever systematic evaluation and benchmarking of state-of-the-art depth of field and defocus deblurring methods across varying focal parameters on non-synthetic, complex, real-world scenes. We demonstrate limitations of current methods: SOTA DoF methods struggle to produce cat-eye vignetting bokeh and non-uniform intensity bokeh, and do not distinguish the focus region clearly, producing similar blur over a range of depth and inconsistent results for nearly identical inputs. SOTA defocus deblur methods demonstrate challenges with sharpening details for wide aperture inputs and tiling artefacts.

\hspace{-1.5em}3) We establish with evidence for the first time that the SOTA DoF methods have non-intuitive parameters that do not tune proportionally to input image focal parameters. Three methods viz. BokehMe, Dr.Bokeh, and BokehDiff show insensitivity to focal length (optimal blur strength parameter does not change with input image focal length). Bokehlicious $\mathcal{F}$-parameter shows a nonlinear trend with input focal length at constant aperture. We highlight the applicability of optically structured, ultra-high-resolution dataset such as MODEST, for rigorous evaluation of SOTA methods and to inform better model architecture designs that include focal parameters as additional inputs for intuitive, proportional tuning.
  
\hspace{-1.5em}4) We release the MODEST dataset, calibration images, data processing and evaluation tools to push research frontiers for non-commercial, academic purposes.

\subsection{Related Datasets}\label{subsec:related_datasets}

%\textcolor{red}{ Some sentences relating and pointing out where prior art lacks w.r.t. our contributions. cat-eye vignetting bokeh and non-uniform intensity bokeh, multiple captures (look at suppl result figure). We want to highlight an additional limitation of prior datasets and profound impact of it on DoF model development. As we show in sections ref(tuning, sens.), the optically structured systematic capture protocol of MODEST enabled us to discover insensitivity of configurable model parameters of these DoF models to the camera capture settings or image attributes. I.e., the optimal value of configurable parameters was found to be either insensitive or nonlinearly changing with capture settings such as focal length, aperture etc. Had the prior datasets been captured with the same systematic focal length and aperture full-range coverage, the DoF model development would have discovered the same findings years ago. This demonstrates the crucical contribution of optically structured capture protocol and full focal coverage of MODEST  in computational photography algorithms development, evaluation, and parameter sensitivity analysis.}
% add one small paragraph to be a summary.
In this subsection, we briefly discuss related existing datasets for depth of field and defocus deblurring mentioned in Table~\ref{table:dataset_comparison}.

% Shallow depth of field rendering

Shallow depth of field (DoF) rendering is a process of synthetically creating the photographic effect where only a small portion of the scene in images is in focus, while the rest is blurred. The idea of shallow DoF is to mimic the optics of real cameras with wide apertures. The CUHK~\citep{Shi2014BlurDetection} dataset is an earlier dataset for DoF applications, but it lacks variations of focal lengths and apertures as images are collected from Internet. DPDD~\citep{abuolaim2020defocus} is a duel-pixel DoF dataset; however, it caters to very specific dual-pixel image sensors with limited disparity because of micro-baselines. DPDD also lacks focal length or aperture variations. EBB!~\citep{ignatov2020rendering} is another DoF dataset consisting low-resolution sharp and blurry image pairs, without image-capture metadata, and without focal length or aperture variations. Recent datasets for shallow DoF rendering are BLB~\citep{bokehme_blb} and VABD~\citep{chen2024variable}. Both datasets solve the issue of single optics and one blur condition in DPDD~\citep{abuolaim2020defocus} by providing either multiple blur levels per scene or capturing multiple apertures per scene. BLB is entirely synthetic, rendered with a computer graphics software and limited camera modelling. Both are limited in resolution as well as number of images. A recent DoF dataset is RealBokeh~\citep{seizinger2025bokehlicious}. While it captures high-resolution RGB images with aperture variations across a large number of scenes, focal lengths are not captured systematically but rather left to the photographer's choice. Empirical assessment showed heavy imbalance of focal lengths favoring narrow FoV or zoomed-in scenes. Furthermore, as RealBokeh prioritizes scene diversity over per-scene optical coverage, it does not capture the same scene across a systematic grid of focal length and aperture combinations, making it difficult to disentangle blur caused by aperture width from blur caused by subject distance. All of the above datasets lack stereo or multi-camera vision, calibration sets, systematic focal length and aperture variations. MODEST addresses all three of these limitations through stereo capture, per-focal-length calibration sets, and a controlled 50 optical configuration grid per scene.

% DoF capture setting discussion.
To capture DoF images with variations in focal lengths, distances, and apertures, a light-field camera can record the spatial location and angular directions of incoming light rays. The images in both RTF~\citep{d2016non} and LFDOF~\citep{ruan2021aifnet} datasets are collected by a light-field camera. RTF contains images captured by physical aperture variations, but LFDOF is captured with a fixed camera aperture. Additionally, for both datasets, the captured images are low resolution due to the nature of the light-field camera. Unlike RTF and LFDOF, RealDOF~\citep{lee2021iterative} is captured by a dual-camera sensor that simulates shallow DoF by retaining sharp subjects in focus and blurring the background; however, a dual-camera has a small baseline between the two cameras and cannot reliably estimate disparities farther away from camera.

The DeepLens~\citep{lijun2018deeplens} and iDFD~\citep{nazir2023idfd} datasets create blurred images through stereo disparity, which is a baseline between a wide-angle lens and a telephoto lens camera, created by a dual-lens camera. Then, it applies the disparity map to compute pixel-wise blurriness levels for foreground and background in the image. Due to the disparity map, the constraints of the dual-lens camera contain boundary confusion and depth Discontinuity. The QPDD~\citep{chen2025quad} dataset contains blurred images captured by a quad-pixel sensor that captures defocus information by splitting the light that enters a single microlens into four different sub-aperture views; however, the blurred images captured by the quad-pixel sensor are difficult to distinguish between actual depth and sensor noise due to the tiny physical distance between the sub-pixels.

% Optical illusions
%Optical illusions in depth-model evaluation refer to probing the limitations of depth-estimation models via perceptual tricks or visually deceptive scenes. These illusions reveal where a model relies on texture, shading, or priors in the images rather than true geometric reasoning. 3D-Visual-Illusion~\citep{yao20253d} is a dataset for optical illusion, and it has been used for human vision experiments where illusions help understand how the human visual system infers 3D structure rather than directly measuring it. The limitation of this dataset is the lack of calibration sets for image calibration. The MonoTrap~\citep{bartolomei2025stereo} is a recent dataset for optical illusions; however, this data only has low-resolution images for the optical illusion task. Thus, many optical illusions in the real world (e.g., heat haze or subtle glass reflection) are missing in low-resolution images. Additionally, the optical illusion images are captured at a fixed number of apertures. This causes MonoTrap to be incapable of mimicking how real professional cameras operate.

% ours (MODEST)
Compared to existing datasets, our proposed MODEST dataset includes both high-resolution stereo RGB images captured with different focal lengths and aperture settings per scene, along with calibration sets provided for calibration algorithms. As calibration algorithms keep eloving, MODEST calibration images enable recalibration or adaptation to lens drift or new alignment. With ultra-high-resolution images (20MP), MODEST enables captured images to preserve much detail for optical fidelity and realistic DoF without the depth blending or information loss of lower resolution datasets.  Additionally, the benefit of MODEST dataset for shallow DoF rendering tasks is that it contains physical defocused and sharp images (i.e., ground truth) due to its wide range of apertures. In addition, in contrast to the RealBokeh dataset, MODEST captures the same scene across a mixture of 50 different optical configurations. This allows researchers to assess the model’s quantitative results through full optical range. Since captured images are obtained from near and distant focal lengths, MODEST also contains cat-eye vignetting and non-uniform intensity bokeh where we show that current SOTA DoF methods fail (supplementary material). MODEST also provides at least two shots from the same viewpoint and exact optical configuration, enabling model assessment for near-identical captures (supplementary material). Unlike prior datasets with limited variations or low resolution, MODEST with its balanced capture is used to  assess senstivity and tuning effects of configurable paramaters in SOTA DoF models. Our study demonstrates insensitivity of configurable model parameters of these DoF models to camera capture settings such as focal length and aperture, that is, the models have non-intuitive tunable parameters that don't change proportionally to image focal lengths or apertures. This finding demonstrates the crucial need of optically structured datasets such as MODEST for computational photography algorithm development, evaluation, and parameter sensitivity analysis.\vspace{-6pt}

\section{Data Acquisition and Processing}\label{sec:method}

\begin{figure}[htbp]  % * makes it span both columns
    %\centering
    \setlength{\abovecaptionskip}{2pt}  % reduce space above caption
    \setlength{\belowcaptionskip}{2pt}  % optional: reduce space below caption   

    \begin{minipage}{\textwidth}
    \centering
    \begin{minipage}[c]{0.22\textwidth}
    \centering
    \includegraphics[width=\textwidth, height=0.8\textwidth]{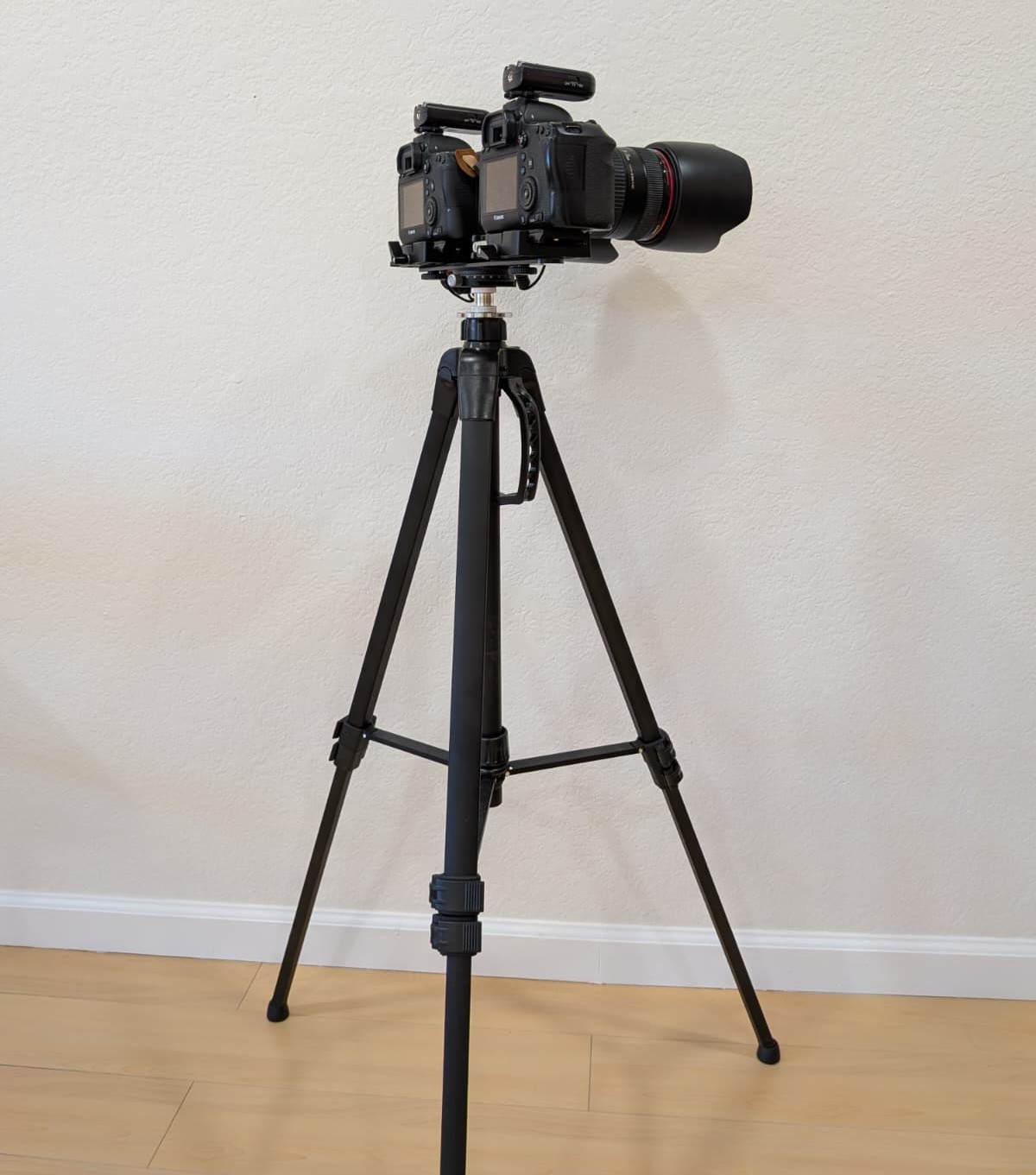}\\[2pt]
    \includegraphics[width=\textwidth, height=0.8\textwidth]{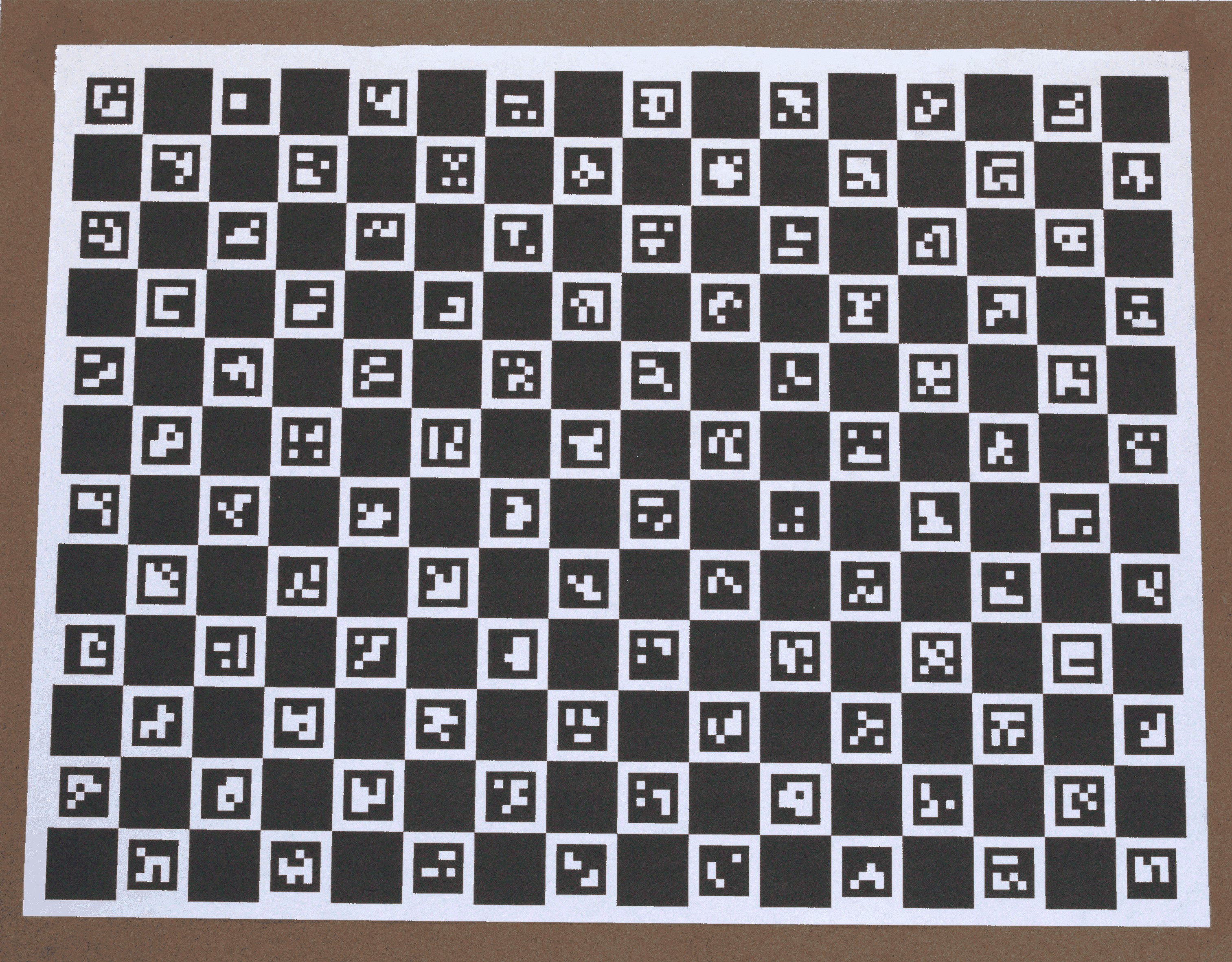}
    \end{minipage}\hfill
    \begin{minipage}[c]{0.77\textwidth}
    \centering
    \includegraphics[width=\textwidth]{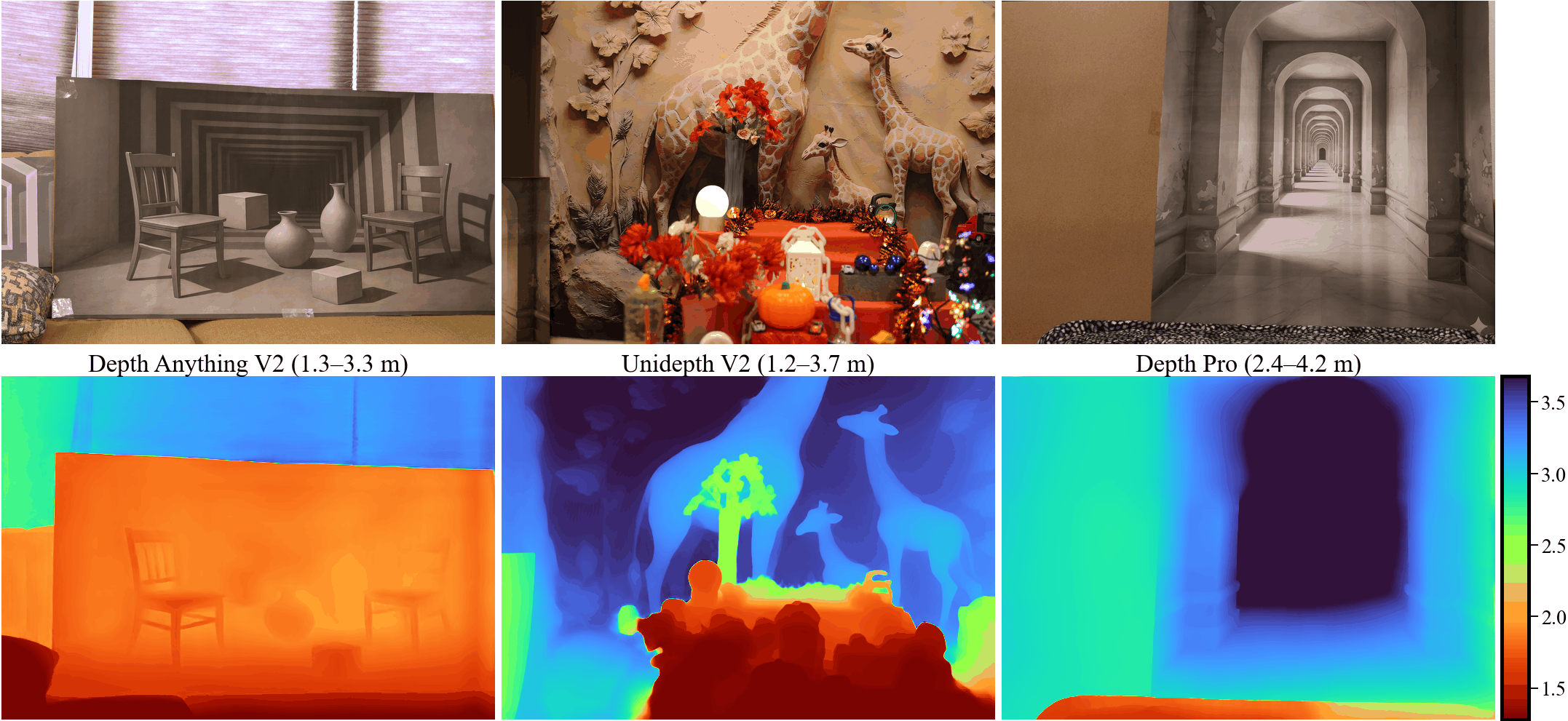}    
    \end{minipage}
    \subcaption*{(a) Stereo camera assembly and ChArUco calibration pattern. (b) Optical illusions where state-of-the-art depth estimators struggle.}\vfil
    
    \begin{minipage}[t]{0.498\textwidth}    
    \includegraphics[width=\textwidth]{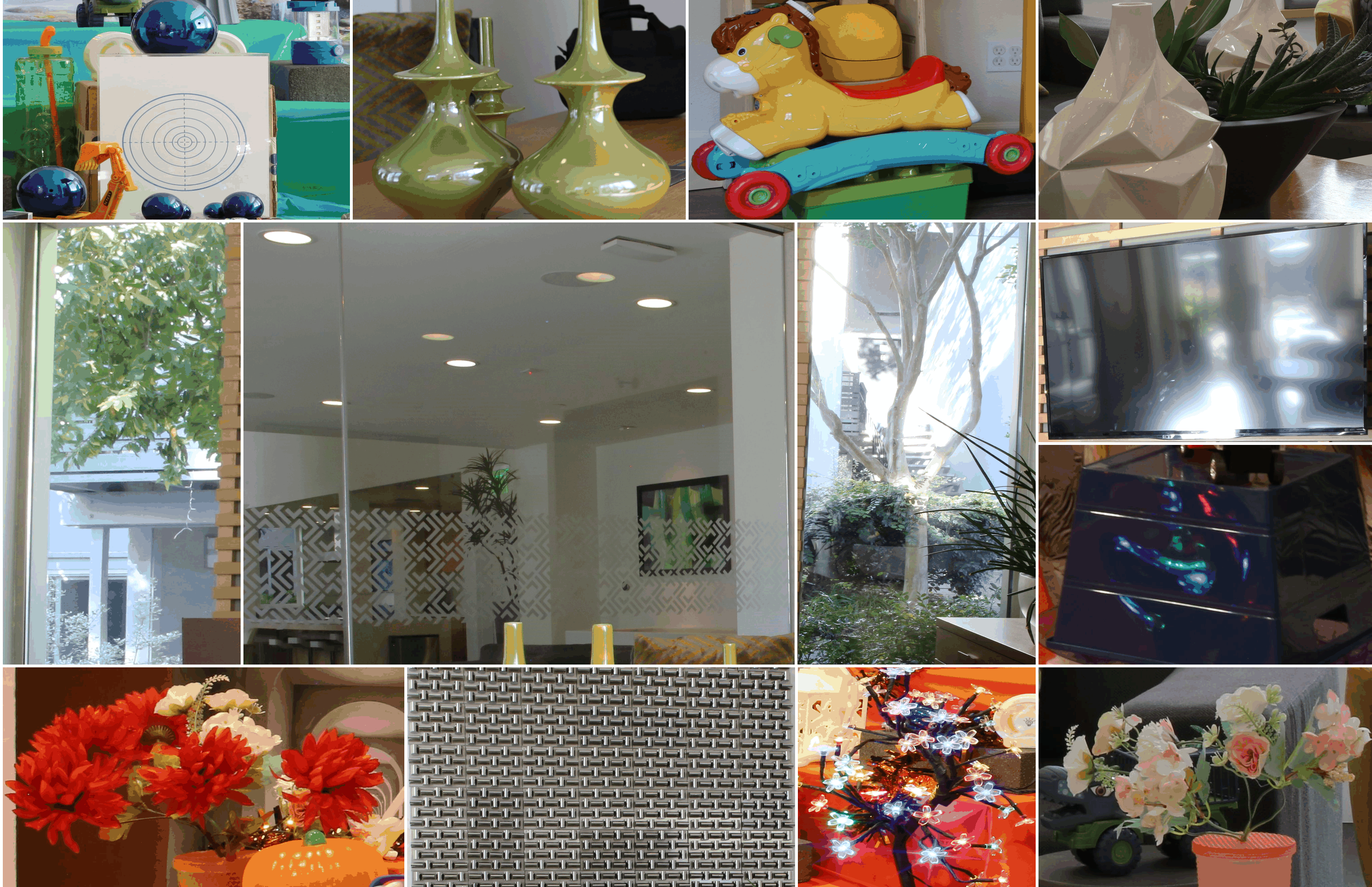}  
    \subcaption*{(c) Challenging scenes containing reflective, fine-detail, semi-transparent, and heterogeneous textured surface}    
    \end{minipage}\hfill
    \begin{minipage}[t]{0.498\textwidth}    
    \includegraphics[width=\textwidth, height=0.64\textwidth]{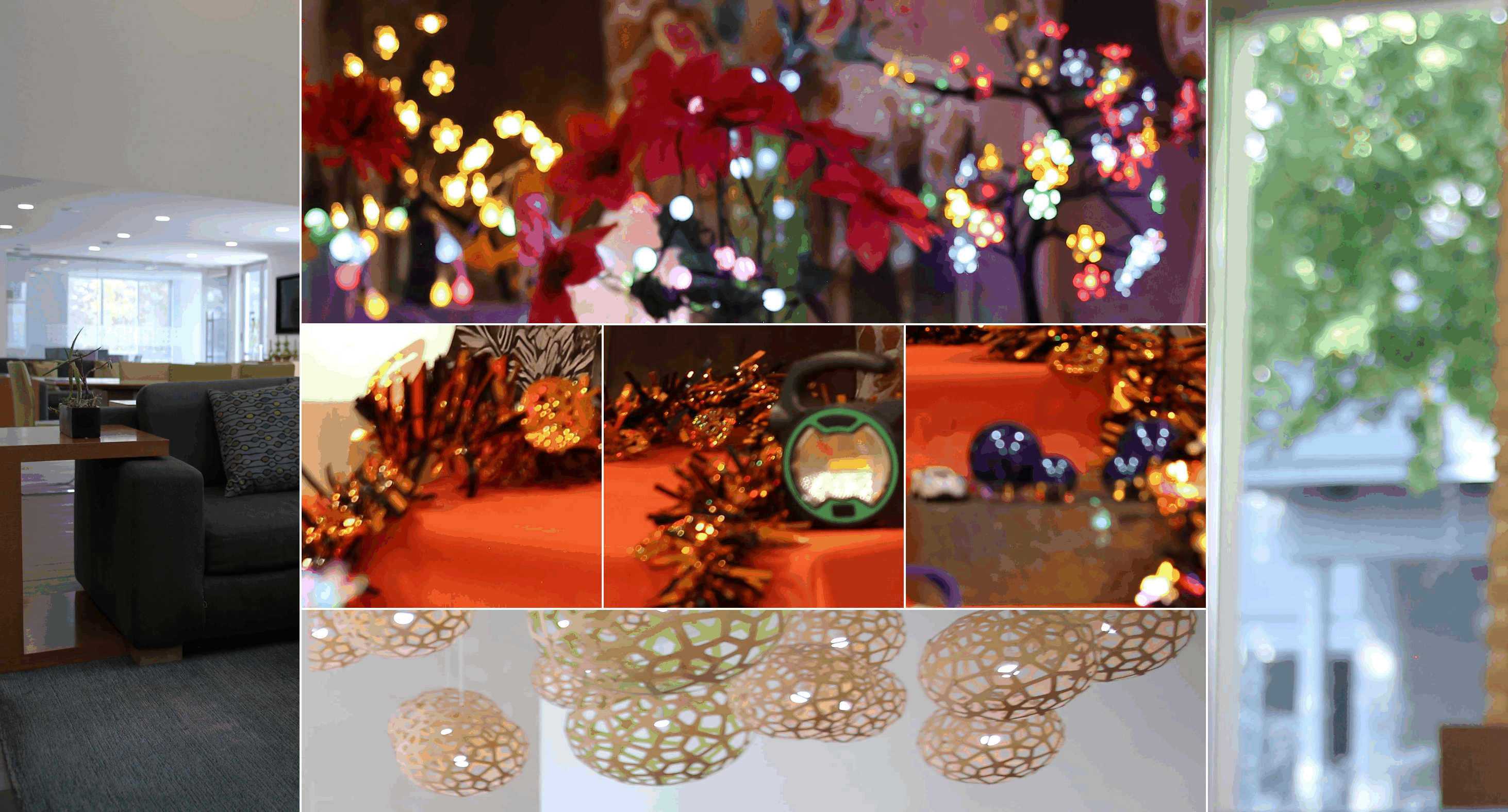}
    \subcaption*{(d) Depth of field effects.}
    \end{minipage}
    
    \includegraphics[width=\textwidth]{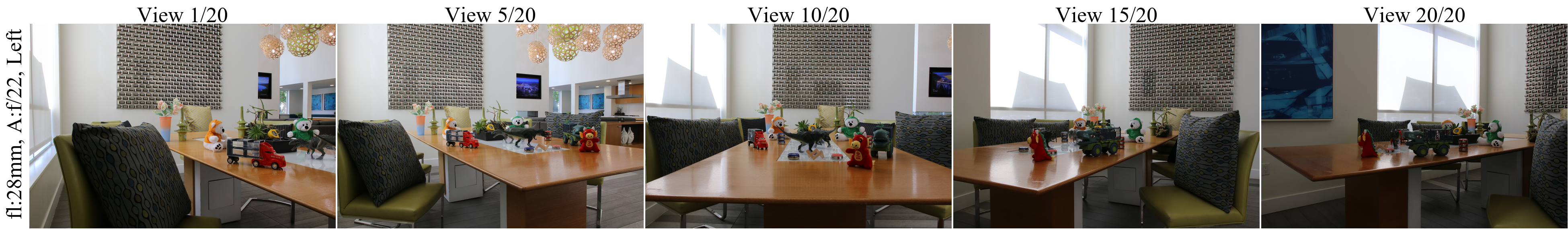}\\
    \includegraphics[width=\textwidth]{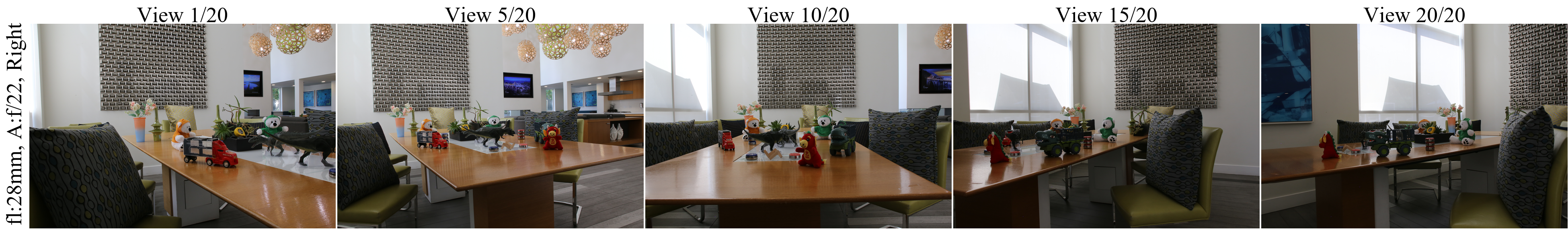}
    \subcaption*{(e) Average 20 viewpoints per focal-length and aperture, with left and right camera per scene.}\vfil
    \includegraphics[width=\textwidth]{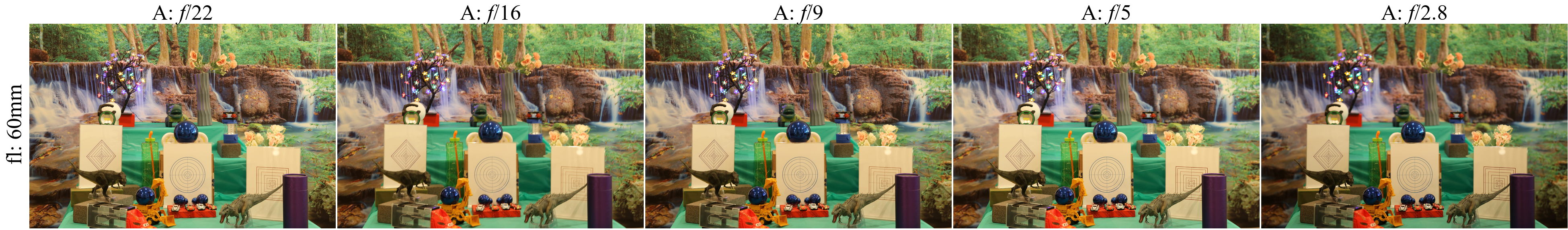}
    \subcaption*{(f) 5 apertures per each of 10 focal lengths per scene.}\vfil
    %{Fig. 2: Data components and structure.}
    \end{minipage}
    
    \caption{MODEST data components and organization.}
    \label{fig:data_comps}
    
\end{figure}

The following sections detail the data collection protocol, calibration process, new challenging elements and structure of the dataset.\vspace{-6pt} 

% \begin{figure*}[h]  % * makes it span both columns
%     \centering
% % calibration illustration.
% % \begin{wrapfigure}{1\textwidth}
% %     \centering
%     \includegraphics[width=\textwidth]{imgs/calib+cam+refl+detl.pdf}
%     \caption{Data acquisition and scene categories. We utilize a stereo camera assembly and diverse calibration viewpoints to capture objects with complex optical properties, such as reflective surfaces and fine details and 3D textures.}
%     \label{fig:cal_illustration}
% \end{figure*}

\subsection{Camera setup and configuration}\label{subsec:cal_illustration}

The camera assembly consists of two \textit{Canon EOS 6D} full-frame DSLR cameras mounted in a synchronized stereo configuration with identical lens setups as seen in Figure~\ref{fig:data_comps} (a). Each camera is equipped with a 28–70\,mm zoom lens, enabling full-range, controlled variation of focal length. The system supports a wide aperture range from f/22.0 to f/2.8, allowing independent adjustment to systematically capture scenes with varying depth of field and blur characteristics. To ensure alignment between the stereo pair, remote shutter triggering is used to achieve synchronous image capture, preventing motion inconsistencies between the two views. The cameras are mounted on a balanced, height-adjustable tripod rig that allows precise control over the stereo baseline distance. This adjustable baseline upto a max of 30 cm enables flexible capture configurations suitable for different scene scales while maintaining geometric consistency. Each shot captures left and right camera views synchronously (stereo system). Every viewpoint gets at least two shots (multi-capture), useful for analysing how models perform for nearly-identical captures (supplementary material). 
%\textcolor{green}{add other papers/datasets also used only one camera}

We intentionally employ identical camera and lens models for the left and right cameras to isolate the influence of make-dependent optical variations, namely, sensor geometry, resolution, lens motion trajectories, aperture and focal range support etc. This enables more controlled left and right captures with systemic focal traversal, and provides a consistent optical reference that enables fair comparison for computational photography methods. Stereo captures are a unique advantage compared to any prior DoF datasets and are directly relevant for proof-of-concept experiments for modern applications such as AR, VR, smartphones, robot-vision. 
\vspace{-6pt}

\subsection{Data Design and Calibration}\label{subsec:data_design}

% \begin{figure*}[h]  % * makes it span both columns
%     \centering
% % calibration illustration.
% % \begin{wrapfigure}{1\textwidth}
% %     \centering
%     \includegraphics[width=\textwidth]{imgs/calib+cam+refl+detl.pdf}
%     \caption{Data acquisition and scene categories. We utilize a stereo camera assembly and diverse calibration viewpoints to capture objects with complex optical properties, such as reflective surfaces and fine details and 3D textures.}
%     \label{fig:cal_illustration}
% \end{figure*}

% \begin{figure}
%     \centering
%     \includegraphics[scale=0.9]{imgs/calib+cam+refl+detl.png}
%     \caption{ADD CAPTION}
%     \label{fig:cal_illustration}
% \end{figure}
We capture 10 distinct scenes using a balanced stereo setup composed of two identical Canon EOS 6D cameras and lenses at full-frame (20MP) resolution. For each scene, we acquire image sets across 10 different focal lengths: \{28, 32, 36, 40, 45, 50, 55, 60, 65, and 70\}mm to cover widest to narrowest field of view, each with different depth of field. Each focal length yields approximately 100 stereo image pairs: 20 pairs per aperture across five aperture settings (f/2.8, f/5.0, f/9.0, f/16.0, f/22.0).

Unlike large-scale, low-resolution, monocular datasets that prioritize scene diversity, MODEST does not intend to replace them for representation learning. Rather, it complements them with  densely sampled, ultra-high resolution, calibratable, stereo optical observations acquired under controlled focal settings spanning broad optical effects spectrum. While the current release contains ten carefully curated scenes with optically challenging elements, we envision MODEST as a benchmark for evaluation, model analysis, calibration-aware learning, and fine-tuning. Future releases will expand the dataset with additional indoor and outdoor scenes while preserving the same acquisition protocol.

% \textcolor{red}{Outdoor scenes with large static objects (e.g., buildings) may not have the same dense arrangement of optically challenging elements, or outdoor scenes with fine details (e.g., leaves, flowers) may not remain static during the full optical coverage and multi-view capture. Depth models also struggle with open-ended depths such as sky and distant horizoris. Therefore, only collecting pairs of sharp and blurry images in potentially moving scenes would not construct a controlled DoF evaluation benchmark.}

MODEST includes a global intrinsics calibration set as well as per-scene extrinsic calibration sets at each focal length. Figure~\ref{fig:data_comps} (a) shows a 16x12 ChArUco pattern with 43mm squares, rigidly attached with $<1$mm surface deviation, used in the global set. The global calibration images are captured with these settings: near/far, 0-180 degrees for the view angle, and camera pitch up,level, or down. 60-100 images per focal length calibrate the standard pinhole intrinsics and $k_1,k_2,p_1,p_2$ distortion per camera. Then, keeping distortion coefficients fixed and optionally finetuning intrinsics, we use a 7x5 checkerboard to calibrate stereo extrinsics per focal length per scene. The stereo extrinsics calibration set, captured at a sharp aperture for each focal length, consists of stereo image pairs (left and right) acquired from multiple viewpoints to ensure robust and accurate geometric calibration. As shown in Tables ~\ref{tab:monocal_eos6d_a} and ~\ref{tab:monocal_eos6d_b} in the supplementary material, the provided ChArUco calibration images in MODEST achieve RMS reprojection errors of lesss than 0.09px per camera, which represents excellent intrinsics calibration for 20MP full frame sensors.  

Beyond enabling accurate image rectification, the provided calibration sets allow dynamic use of MODEST, as calibration algorithms evolve. Calibration is a crucial step in most computational photography and geometry-based vision tasks and very often lacking from public datasets, compromising benchmarking quality and method potential. We captured calibration sets with future in sight: developing more advanced calibration-aware computational photography methods: geometrically accurate stereo depth supervision through calibrated multi-view reconstruction, physically grounded defocus and bokeh rendering using camera intrinsics, evaluate calibration-aware neural depth estimation methods that explicitly incorporate camera models, investigate focal-length-dependent geometric consistency across multiple optical configurations, and benchmark calibration-sensitive tasks such as novel-view synthesis, Gaussian Splatting, NeRF reconstruction, and physics-based image formation.
\vspace{-6pt}

\subsection{Scene Diversity and Visual Complexity}\label{subsec:data_illu}

\begin{table}[htbp]
\centering
\caption{Types of scene objects included to capture diverse visual and optical properties.}
\label{tab:scene_objects}
\small 
\begin{tabularx}{\linewidth}{>{\raggedright\arraybackslash}p{3cm} X}
\toprule
\textbf{Category} & \textbf{Examples of Objects} \\
\midrule
\textbf{Reflective surfaces} & Reflective spheres, ceramic jars, transparent containers, semi-transparent doors \\ %& Mirrors, reflective spheres, metallic balls, reflective cubes, films, crystal spheres, balloons, glass jars, transparent containers, golden foils. \\
\addlinespace[0.6em]
\textbf{Fine Details} & Small toys, geometric props, stacked objects, patterned surfaces, small structural shapes, flowers, textured objects, shaped lights. \\ %, torch. \\
%\addlinespace[0.6em]
%\textbf{Optical Illusions} & Printed illusion patterns, structured geometric illusion prints, target prints. \\  %& Printed illusion patterns, checkerboard targets, structured geometric illusion prints, target prints. \\
\bottomrule
\end{tabularx}
\end{table}

% --- new text ----
MODEST comprises 10 scenes as illustrated in Figure~\ref{fig:data}, designed to capture diverse visual and optical properties, summarized in Table~\ref{tab:scene_objects}. Scenes 1, 8, and 10 represent dense, complex configuration and are intentionally designed to be highly challenging for DoF and defocus deblurring tasks. These scenes include closely arranged objects, including reflective materials, decorative elements, lamps, and light-emitting sources, along with flowers containing thin petals and stems under varying illumination conditions. These objects are densely arranged to create occlusions and complex depth boundaries, with large 2D depth illusion prints. Scenes 3-7,9 introduce living environments with varying background depth, lighting, and object arrangements: toys, illusion prints, reflective surfaces, semi-transparent glass, natural window lighting.

As shown in Fig. ~\ref{fig:data_comps} (b), state-of-the-art depth estimation models, including Depth Anthing~\citep{yang2024depth}, Unidepth V2~\citep{piccinelli2025unidepthv2}, and Depth Pro~\citep{bochkovskii2024depth}, struggle to accurately estimate depth for the optical illusion patterns in MODEST. Fig.~\ref{fig:data_comps} (c) highlights the diverse visual challenges within the dataset: reflective and semi-transparent surfaces like ceramic containers, TVs, and glass windows, where background elements (e.g., foliage) further complicate depth perception. Some scenes feature fine details including flowers, light-emitting sources, artistic patterns. These objects increase the visual complexity of the scene and task challenges for DoF and deblurring algorithms. Fig.~\ref{fig:data_comps} (d) demonstrates some examples of depth of field effects. The leftmost and rightmost panels showcase focus transitions on semi-transparent surfaces, ranging from sharp foreground sofa and window frames to soft, blurry tree leaves. Central panels illustrate natural optical effects such as varying bokeh shapes and non-uniform color intensity. These variations highlight depth of field effects between different aperture sizes, focal lengths, and subject distances.

Fig.~\ref{fig:data_comps} (e) illustrates a viewpoint shift across a table (views 1 to 20), present in every MODEST scene. The perspective changes reveal different angles of the objects on the table and the background. As the camera moves, different objects become visible or occluded. In this example, focal length of 28mm with an aperture of f/22.0 keeps both the foreground pillows and the background textured piece relatively sharp. Fig.~\ref{fig:data_comps} (f) shows the effect of different apertures using a constant focal length of 60mm. It displays five side-by-side shots with the same view but different apertures. With a small aperture (i.e., f/22.0), the entire scene from the foreground dinosaurs to the waterfall backdrop is in relatively sharp focus. As the f-number decreases, the background becomes blurrier. When a large aperture (i.e., f/2.8 is used, the background waterfall is out of focus, making the foreground objects stand out.

As shown in Table~\ref{tab:multi_capture}, MODEST offers significant advantages over existing datasets like RealBokeh through its systematic optical configuration changes. While both datasets share a similar focal range (28mm–70mm), MODEST uniquely features systematic focal length and aperture variations, and multiple view-points per scene. Furthermore, MODEST is the only dataset to provide stereo images for better disparity or depth map estimation. Its multi-capture capability, capturing the same scene at least twice under identical optical settings, allows for the development and testing of depth of field models that account for subtle lighting variations, tesing robustness against micro input perturbations. MODEST also provides intrinsic and extrinsic calibration images for accurate calibration operation.\vspace{-6pt}

\section{Experiments and Application}

\begin{table*}[htbp]
\centering
\caption{
Structured benchmarking of depth of field methods across five focal lengths for four scenes (S1, S3, S5, S8) with overall average, with the optimal parameter tuning. Higher PSNR/SSIM and lower LPIPS indicate better performance.\vspace{-10pt}}
\resizebox{\textwidth}{!}{
\begin{tabular}{l|c c c c c|c c c c c|c c c c c|c c c}
\toprule
\multirow{2}{*}{\textbf{Method}} &
\multicolumn{5}{c|}{\textbf{PSNR} $\uparrow$} &
\multicolumn{5}{c|}{\textbf{SSIM} $\uparrow$} &
\multicolumn{5}{c|}{\textbf{LPIPS} $\downarrow$} &
\multicolumn{3}{c}{\textbf{Average}} \\
& fl28 & fl36 & fl45 & fl60 & fl70 &
fl28 & fl36 & fl45 & fl60 & fl70 &
fl28 & fl36 & fl45 & fl60 & fl70 &
PSNR$\uparrow$ & SSIM$\uparrow$ & LPIPS$\downarrow$ \\
\midrule
%% ── Scene S1 ──
\multicolumn{19}{l}{\textit{Scene 1}} \\
\midrule
BokehMe &
26.13 & 26.92 & 25.40 & 25.54 & \textcolor{darkgreen}{25.78} &
0.89 & 0.89 & 0.87 & 0.89 & \textcolor{darkgreen}{0.89} &
0.24 & 0.24 & 0.26 & 0.25 & \textcolor{darkgreen}{0.26} &
25.95 & 0.89 & 0.25 \\
Dr.Bokeh &
27.34 & 28.94 & 27.15 & \textcolor{darkgreen}{26.41} & 24.82 &
\textcolor{darkgreen}{0.93} & \textcolor{darkgreen}{0.94} & \textcolor{darkgreen}{0.91} & \textcolor{darkgreen}{0.91} & 0.87 &
0.18 & 0.16 & \textcolor{darkgreen}{0.19} & \textcolor{darkgreen}{0.22} & 0.27 &
26.93 & \textcolor{darkgreen}{0.91} & \textcolor{darkgreen}{0.20} \\
BokehDiff &
24.92 & 27.66 & 23.98 & 21.54 & 20.49 &
0.83 & 0.88 & 0.79 & 0.73 & 0.70 &
0.19 & 0.16 & 0.22 & 0.31 & 0.36 &
23.72 & 0.79 & 0.25 \\
Bokehlicious &
\textcolor{darkgreen}{28.30} & \textcolor{darkgreen}{29.92} & \textcolor{darkgreen}{27.55} & 26.23 & 25.62 &
\textcolor{darkgreen}{0.93} & \textcolor{darkgreen}{0.94} & \textcolor{darkgreen}{0.91} & 0.89 & 0.88 &
\textcolor{darkgreen}{0.14} & \textcolor{darkgreen}{0.14} & \textcolor{darkgreen}{0.19} & 0.27 & \textcolor{darkgreen}{0.26} &
\textcolor{darkgreen}{27.53} & \textcolor{darkgreen}{0.91} & \textcolor{darkgreen}{0.20} \\
\midrule
%% ── Scene S3 ──
\multicolumn{19}{l}{\textit{Scene 3}} \\
\midrule
BokehMe &
27.94 & \textcolor{darkgreen}{28.38} & \textcolor{darkgreen}{27.53} & 24.28 & 24.26 &
0.93 & \textcolor{darkgreen}{0.93} & \textcolor{darkgreen}{0.92} & 0.85 & 0.83 &
0.21 & \textcolor{darkgreen}{0.22} & \textcolor{darkgreen}{0.25} & 0.34 & 0.35 &
26.48 & \textcolor{darkgreen}{0.90} & \textcolor{darkgreen}{0.27} \\
Dr.Bokeh &
\textcolor{darkgreen}{28.46} & 28.15 & 26.68 & 23.65 & 24.23 &
\textcolor{darkgreen}{0.94} & 0.92 & 0.90 & 0.82 & 0.82 &
\textcolor{darkgreen}{0.20} & 0.23 & 0.27 & 0.37 & 0.36 &
26.24 & 0.88 & 0.29 \\
BokehDiff &
28.05 & 27.67 & 25.86 & 23.13 & 23.98 &
0.92 & 0.90 & 0.87 & 0.79 & 0.80 &
0.21 & 0.24 & 0.29 & 0.38 & 0.37 &
25.74 & 0.86 & 0.30 \\
Bokehlicious &
28.28 & 27.86 & 26.68 & \textcolor{darkgreen}{25.01} & \textcolor{darkgreen}{24.90} &
0.93 & 0.91 & 0.89 & \textcolor{darkgreen}{0.88} & \textcolor{darkgreen}{0.84} &
\textcolor{darkgreen}{0.20} & 0.23 & 0.27 & \textcolor{darkgreen}{0.30} & \textcolor{darkgreen}{0.33} &
\textcolor{darkgreen}{26.55} & 0.89 & \textcolor{darkgreen}{0.27} \\
\midrule
%% ── Scene S5 ──
\multicolumn{19}{l}{\textit{Scene 5}} \\
\midrule
BokehMe &
28.14 & 29.26 & 27.53 & 26.04 & 26.45 &
0.91 & 0.91 & 0.89 & 0.86 & 0.87 &
0.25 & 0.28 & 0.32 & 0.36 & 0.36 &
27.48 & 0.89 & 0.31 \\
Dr.Bokeh &
29.46 & 29.74 & 28.67 & 26.20 & 26.81 &
\textcolor{darkgreen}{0.94} & \textcolor{darkgreen}{0.92} & 0.91 & 0.86 & 0.87 &
0.20 & 0.25 & 0.27 & 0.35 & 0.35 &
28.17 & 0.90 & 0.28 \\
BokehDiff &
29.43 & 30.16 & 28.64 & 26.14 & 26.83 &
0.92 & 0.91 & 0.90 & 0.85 & 0.86 &
0.20 & 0.26 & 0.27 & 0.36 & 0.36 &
28.24 & 0.89 & 0.29 \\
Bokehlicious &
\textcolor{darkgreen}{29.73} & \textcolor{darkgreen}{30.42} & \textcolor{darkgreen}{29.22} & \textcolor{darkgreen}{27.78} & \textcolor{darkgreen}{28.02} &
\textcolor{darkgreen}{0.94} & \textcolor{darkgreen}{0.92} & \textcolor{darkgreen}{0.92} & \textcolor{darkgreen}{0.91} & \textcolor{darkgreen}{0.91} &
\textcolor{darkgreen}{0.18} & \textcolor{darkgreen}{0.22} & \textcolor{darkgreen}{0.23} & \textcolor{darkgreen}{0.27} & \textcolor{darkgreen}{0.28} &
\textcolor{darkgreen}{29.03} & \textcolor{darkgreen}{0.92} & \textcolor{darkgreen}{0.23} \\
\midrule
%% ── Scene S8 ──
\multicolumn{19}{l}{\textit{Scene 8}} \\
\midrule
BokehMe &
27.11 & 28.62 & 26.29 & 26.00 & 26.01 &
0.88 & \textcolor{darkgreen}{0.90} & 0.83 & 0.85 & 0.82 &
0.27 & 0.22 & 0.33 & 0.30 & 0.35 &
26.81 & 0.86 & 0.29 \\
Dr.Bokeh &
\textcolor{darkgreen}{28.26} & \textcolor{darkgreen}{28.63} & 27.52 & 26.59 & 26.35 &
\textcolor{darkgreen}{0.91} & 0.89 & 0.86 & 0.87 & 0.84 &
0.21 & 0.23 & 0.27 & 0.27 & 0.32 &
27.47 & 0.87 & 0.26 \\
BokehDiff &
28.04 & 27.95 & 26.87 & 26.36 & 26.30 &
0.89 & 0.87 & 0.83 & 0.84 & 0.82 &
0.20 & 0.22 & 0.27 & 0.27 & 0.31 &
27.11 & 0.85 & 0.25 \\
Bokehlicious &
28.17 & 28.17 & \textcolor{darkgreen}{27.82} & \textcolor{darkgreen}{26.92} & \textcolor{darkgreen}{27.45} &
0.89 & 0.88 & \textcolor{darkgreen}{0.87} & \textcolor{darkgreen}{0.89} & \textcolor{darkgreen}{0.89} &
\textcolor{darkgreen}{0.19} & \textcolor{darkgreen}{0.21} & \textcolor{darkgreen}{0.25} & \textcolor{darkgreen}{0.24} & \textcolor{darkgreen}{0.27} &
\textcolor{darkgreen}{27.71} & \textcolor{darkgreen}{0.88} & \textcolor{darkgreen}{0.23} \\
\midrule
%% ── 4-Scene Average ──
\multicolumn{19}{l}{\textit{4-Scene Average}} \\
\midrule
BokehMe &
27.33 & 28.29 & 26.69 & 25.47 & 25.62 &
0.90 & 0.91 & 0.88 & 0.87 & 0.86 &
0.24 & 0.24 & 0.29 & 0.31 & 0.33 &
26.68 & 0.88 & 0.28 \\
Dr.Bokeh &
28.38 & 28.86 & 27.51 & 25.71 & 25.55 &
\textcolor{darkgreen}{0.93} & \textcolor{darkgreen}{0.92} & \textcolor{darkgreen}{0.90} & 0.87 & 0.85 &
0.20 & 0.22 & 0.25 & 0.30 & 0.32 &
27.20 & 0.89 & 0.26 \\
BokehDiff &
27.61 & 28.36 & 26.34 & 24.29 & 24.40 &
0.89 & 0.89 & 0.85 & 0.80 & 0.79 &
0.20 & 0.22 & 0.26 & 0.33 & 0.35 &
26.20 & 0.84 & 0.27 \\
Bokehlicious & 
\textcolor{darkgreen}{28.62} & \textcolor{darkgreen}{29.09} & \textcolor{darkgreen}{27.82} & \textcolor{darkgreen}{26.48} & \textcolor{darkgreen}{27.70} &
0.92 & 0.91 & \textcolor{darkgreen}{0.90} & \textcolor{darkgreen}{0.89} & \textcolor{darkgreen}{0.90} &
\textcolor{darkgreen}{0.18} & \textcolor{darkgreen}{0.20} & \textcolor{darkgreen}{0.23} & \textcolor{darkgreen}{0.27} & \textcolor{darkgreen}{0.23} &
\textcolor{darkgreen}{28.01} & \textcolor{darkgreen}{0.90} & \textcolor{darkgreen}{0.22} \\
\bottomrule
\end{tabular}}
\label{tab:dof_results}
\vspace{-7pt}
\end{table*}

\begin{table}[htbp]
\centering
\caption{
Mean-opinion-rank (MOR) over 530 rankings and inference execution time(s) per image on A100-40GB GPU for four depth of field methods. Lower MOR and execution time indicate better performance. \vspace{-10pt}}
\resizebox{0.5\textwidth}{!}{
\begin{tabular}{c|c c}
\toprule
\textbf{Method} &MOR$\downarrow$  &Time(s)$\downarrow$\\
\midrule
BokehMe ~\citep{peng2022bokehme} &\textcolor{darkgreen}{2.32}  &17.7+12.2 \\%
Dr.Bokeh ~\citep{sheng2024dr} &2.54 &120.4 \\%
BokehDiff ~\citep{zhu2025bokehdiff} &2.80 & 23.1+2e-6 \\%
Bokehlicious  ~\citep{seizinger2025bokehlicious} &2.35 &\textcolor{darkgreen}{17.8}\\
\bottomrule
\end{tabular}}
\label{tab:dof_mor}
\vspace{-10pt}
\end{table}

\begin{table}[htbp]
\centering
\caption{
Quantitative evaluation of depth of field models on Scene~1, using \textit{stereo depth} predicted by FoundationStereo~\citep{foundation}  as input, across five focal lengths. Compare with Scene1 monocular results in Tab.~\ref{tab:dof_results}. 
\vspace{-10pt}}
\resizebox{\textwidth}{!}{
\begin{tabular}{l|c c c c c|c c c c c|c c c c c|c c c}
\toprule
\multirow{2}{*}{\textbf{Method}} &
\multicolumn{5}{c|}{\textbf{PSNR} $\uparrow$} &
\multicolumn{5}{c|}{\textbf{SSIM} $\uparrow$} &
\multicolumn{5}{c|}{\textbf{LPIPS} $\downarrow$} &
\multicolumn{3}{c}{\textbf{Average}} \\
& fl28 & fl36 & fl45 & fl60 & fl70 &
fl28 & fl36 & fl45 & fl60 & fl70 &
fl28 & fl36 & fl45 & fl60 & fl70 &
PSNR$\uparrow$ & SSIM$\uparrow$ & LPIPS$\downarrow$ \\
\midrule
%% ────────────────────────── STEREO-DEPTH INPUT (FoundationStereo), SCENE 1 ──────────────────────────
\multicolumn{19}{l}{\textit{Scene 1 stereo depth}} \\
\midrule
BokehMe &
26.90 & 27.62 & 26.22 & 25.82 & 25.52 &
0.89 & 0.89 & 0.88 & \textcolor{darkgreen}{0.88} & 0.88 &
0.23 & 0.22 & 0.23 & 0.24 & 0.27 &
26.41 & 0.89 & 0.24 \\
Dr.Bokeh &
28.78 & 29.61 & 27.68 & 25.95 & 25.02 &
\textcolor{darkgreen}{0.93} & \textcolor{darkgreen}{0.93} & \textcolor{darkgreen}{0.90} & \textcolor{darkgreen}{0.88} & 0.87 &
0.17 & 0.17 & \textcolor{darkgreen}{0.19} & \textcolor{darkgreen}{0.22} & 0.28 &
27.41 & \textcolor{darkgreen}{0.90} & \textcolor{darkgreen}{0.21} \\
BokehDiff &
28.49 & 29.40 & 27.17 & 25.98 & 25.07 &
0.89 & 0.89 & 0.87 & 0.87 & 0.85 &
0.25 & 0.24 & 0.26 & 0.26 & 0.30 &
27.22 & 0.88 & 0.26 \\
Bokehlicious &
\textcolor{darkgreen}{29.99} & \textcolor{darkgreen}{31.23} & \textcolor{darkgreen}{27.92} & \textcolor{darkgreen}{26.07} & \textcolor{darkgreen}{26.26} &
\textcolor{darkgreen}{0.93} & \textcolor{darkgreen}{0.93} & \textcolor{darkgreen}{0.90} & 0.87 & \textcolor{darkgreen}{0.89} &
\textcolor{darkgreen}{0.14} & \textcolor{darkgreen}{0.15} & 0.20 & 0.28 & \textcolor{darkgreen}{0.26} &
\textcolor{darkgreen}{28.29} & \textcolor{darkgreen}{0.90} & \textcolor{darkgreen}{0.21} \\
\bottomrule
\end{tabular}}
\label{tab:stereo_dof_results}
\vspace{-10pt}
\end{table}

\begin{table}[htbp]
\centering
\caption{
Quantitative evaluation of defocus deblurring methods across five focal lengths and the overall average for four scenes (S1, S3, S5, S8), including inference execution time(s) per image. Higher PSNR/SSIM and lower LPIPS indicate better performance. Two motion-deblurring methods are stress-tested on defocus deblurring.\vspace{-10pt}}
\resizebox{\textwidth}{!}{
\begin{tabular}{l|c c c c c|c c c c c|c c c c c|c c c c}
\toprule
\multirow{2}{*}{\textbf{Method}} &
\multicolumn{5}{c|}{\textbf{PSNR} $\uparrow$} &
\multicolumn{5}{c|}{\textbf{SSIM} $\uparrow$} &
\multicolumn{5}{c|}{\textbf{LPIPS} $\downarrow$} &
\multicolumn{4}{c}{\textbf{Average}} \\
& fl28 & fl36 & fl45 & fl60 & fl70 &
fl28 & fl36 & fl45 & fl60 & fl70 &
fl28 & fl36 & fl45 & fl60 & fl70 &
PSNR$\uparrow$ & SSIM$\uparrow$ & LPIPS$\downarrow$ & Time(s)$\downarrow$ \\
\midrule
%% ────────────────────────── DEFOCUS DEBLURRING ──────────────────────────
\multicolumn{20}{l}{\textbf{Defocus Deblurring Methods}} \\
\midrule
%% ── Scene S1 ──
\multicolumn{20}{l}{\textit{Scene 1}} \\
\midrule
Restormer (`22)~\citep{zamir2022restormer} &
27.19 & 28.51 & 26.99 & 26.42 & 25.62 &
\textcolor{darkgreen}{0.91} & \textcolor{darkgreen}{0.91} & \textcolor{darkgreen}{0.89} & \textcolor{darkgreen}{0.90} & \textcolor{darkgreen}{0.87} &
\textcolor{darkgreen}{0.03} & \textcolor{darkgreen}{0.03} & \textcolor{darkgreen}{0.05} & \textcolor{darkgreen}{0.06} & \textcolor{darkgreen}{0.08} &
26.95 & \textcolor{darkgreen}{0.90} & \textcolor{darkgreen}{0.05} & 6.36 \\
NRKNet (`24)~\citep{quan2023neumann} &
\textcolor{darkgreen}{27.50} & \textcolor{darkgreen}{28.97} & \textcolor{darkgreen}{27.30} & 26.41 & 25.53 &
\textcolor{darkgreen}{0.91} & \textcolor{darkgreen}{0.91} & \textcolor{darkgreen}{0.89} & 0.89 & \textcolor{darkgreen}{0.87} &
0.04 & \textcolor{darkgreen}{0.03} & 0.06 & 0.08 & 0.11 &
\textcolor{darkgreen}{26.98} & 0.89 & 0.07 & \textcolor{darkgreen}{0.35} \\
ViTDeblur (`24)~\citep{liang2024vitdeblur} &
26.83 & 27.94 & 27.00 & \textcolor{darkgreen}{26.54} & \textcolor{darkgreen}{25.79} &
\textcolor{darkgreen}{0.91} & \textcolor{darkgreen}{0.91} & \textcolor{darkgreen}{0.89} & \textcolor{darkgreen}{0.90} & \textcolor{darkgreen}{0.87} &
0.04 & \textcolor{darkgreen}{0.03} & \textcolor{darkgreen}{0.05} & \textcolor{darkgreen}{0.06} & \textcolor{darkgreen}{0.08} &
26.73 & \textcolor{darkgreen}{0.90} & \textcolor{darkgreen}{0.05} & 12.91 \\
\midrule
%% ── Scene S3 ──
\multicolumn{20}{l}{\textit{Scene 3}} \\
\midrule
Restormer (`22)~\citep{zamir2022restormer} &
26.28 & 25.53 & 25.74 & 24.16 & 25.35 &
{0.89} & {0.87} & \textcolor{darkgreen}{0.86} & \textcolor{darkgreen}{0.84} & \textcolor{darkgreen}{0.85} &
\textcolor{darkgreen}{0.07} & \textcolor{darkgreen}{0.09} & 0.11 & \textcolor{darkgreen}{0.15} & \textcolor{darkgreen}{0.11} &
25.45 & \textcolor{darkgreen}{0.86} & \textcolor{darkgreen}{0.10} & 6.36 \\
NRKNet (`24)~\citep{quan2023neumann} &
\textcolor{darkgreen}{26.69} & \textcolor{darkgreen}{26.87} & 25.93 & 24.28 & 25.17 &
\textcolor{darkgreen}{0.90} & \textcolor{darkgreen}{0.88} & 0.85 & 0.83 & 0.83 &
\textcolor{darkgreen}{0.07} & \textcolor{darkgreen}{0.09} & 0.13 & 0.19 & 0.17 &
25.87 & \textcolor{darkgreen}{0.86} & 0.13 & \textcolor{darkgreen}{0.35} \\
ViTDeblur (`24)~\citep{liang2024vitdeblur} &
26.58 & 26.58 & \textcolor{darkgreen}{26.13} & \textcolor{darkgreen}{24.48} & \textcolor{darkgreen}{25.93} &
0.89 & \textcolor{darkgreen}{0.88} & \textcolor{darkgreen}{0.86} & \textcolor{darkgreen}{0.84} & \textcolor{darkgreen}{0.85} &
\textcolor{darkgreen}{0.07} & \textcolor{darkgreen}{0.09} & \textcolor{darkgreen}{0.10} & \textcolor{darkgreen}{0.15} & {0.13} &
\textcolor{darkgreen}{26.01} & \textcolor{darkgreen}{0.86} & {0.11} & 12.91 \\
\midrule
%% ── Scene S5 ──
\multicolumn{20}{l}{\textit{Scene 5}} \\
\midrule
Restormer (`22)~\citep{zamir2022restormer} &
\textcolor{darkgreen}{29.03} & 29.90 & 29.07 & \textcolor{darkgreen}{27.81} & \textcolor{darkgreen}{28.66} &
\textcolor{darkgreen}{0.92} & \textcolor{darkgreen}{0.91} & \textcolor{darkgreen}{0.90} & \textcolor{darkgreen}{0.89} & \textcolor{darkgreen}{0.91} &
\textcolor{darkgreen}{0.05} & \textcolor{darkgreen}{0.05} & \textcolor{darkgreen}{0.05} & \textcolor{darkgreen}{0.09} & \textcolor{darkgreen}{0.09} &
28.89 & \textcolor{darkgreen}{0.91} & \textcolor{darkgreen}{0.07} & 6.36 \\
NRKNet (`24)~\citep{quan2023neumann} &
28.85 & {30.07} & 28.86 & 27.27 & 27.90 &
\textcolor{darkgreen}{0.92} & \textcolor{darkgreen}{0.91} & 0.89 & 0.88 & 0.90 &
\textcolor{darkgreen}{0.05} & 0.07 & 0.08 & 0.15 & 0.17 &
28.57 & 0.90 & 0.11 & \textcolor{darkgreen}{0.35} \\
ViTDeblur (`24)~\citep{liang2024vitdeblur} &
{28.96} & \textcolor{darkgreen}{30.35} & \textcolor{darkgreen}{29.36} & \textcolor{darkgreen}{27.83} & 28.33 &
\textcolor{darkgreen}{0.92} & \textcolor{darkgreen}{0.91} & \textcolor{darkgreen}{0.90} & \textcolor{darkgreen}{0.89} & \textcolor{darkgreen}{0.91} &
\textcolor{darkgreen}{0.05} & \textcolor{darkgreen}{0.05} &{0.06} & {0.11} & {0.11} &
\textcolor{darkgreen}{28.95} & \textcolor{darkgreen}{0.91} & \textcolor{darkgreen}{0.08} & 12.91 \\
\midrule
%% ── Scene S8 ──
\multicolumn{20}{l}{\textit{Scene 8}} \\
\midrule
Restormer (`22)~\citep{zamir2022restormer} &
26.54 & 26.09 & 26.40 & \textcolor{darkgreen}{26.79} & 26.72 &
\textcolor{darkgreen}{0.87} & {0.84} & \textcolor{darkgreen}{0.83} & \textcolor{darkgreen}{0.87} & \textcolor{darkgreen}{0.86} &
\textcolor{darkgreen}{0.05} & \textcolor{darkgreen}{0.06} & \textcolor{darkgreen}{0.10} & \textcolor{darkgreen}{0.08} & \textcolor{darkgreen}{0.10} &
26.52 & \textcolor{darkgreen}{0.85} & \textcolor{darkgreen}{0.08} & 6.36 \\
NRKNet (`24)~\citep{quan2023neumann} &
\textcolor{darkgreen}{27.38} & \textcolor{darkgreen}{27.56} & \textcolor{darkgreen}{26.75} & 26.67 & 26.72 &
\textcolor{darkgreen}{0.87} & \textcolor{darkgreen}{0.85} & 0.82 & {0.86} & \textcolor{darkgreen}{0.86} &
0.06 & 0.07 & 0.11 & 0.12 & 0.14 &
\textcolor{darkgreen}{26.99} & \textcolor{darkgreen}{0.85} & 0.11 & \textcolor{darkgreen}{0.35} \\
ViTDeblur (`24)~\citep{liang2024vitdeblur} &
26.51 & 26.60 & 26.60 & 26.64 & 26.75 &
0.86 & \textcolor{darkgreen}{0.85} & \textcolor{darkgreen}{0.83} & {0.86} & \textcolor{darkgreen}{0.86} &
0.06 & 0.07 & \textcolor{darkgreen}{0.10} & {0.09} & {0.11} &
26.62 & \textcolor{darkgreen}{0.85} & \textcolor{darkgreen}{0.09} & 12.91 \\
\midrule

\midrule
%% ── 4-Scene Average ──
\multicolumn{20}{l}{\textit{4-Scene Average}} \\
\midrule
Restormer (`22)~\citep{zamir2022restormer} &
27.26 & 27.51 & 27.05 & 26.30 & 26.59 &
0.90 & 0.88 & \textcolor{darkgreen}{0.87} & \textcolor{darkgreen}{0.88} & \textcolor{darkgreen}{0.87} &
\textcolor{darkgreen}{0.05} & \textcolor{darkgreen}{0.06} & \textcolor{darkgreen}{0.08} & \textcolor{darkgreen}{0.10} & \textcolor{darkgreen}{0.10} &
26.94 & \textcolor{darkgreen}{0.88} & \textcolor{darkgreen}{0.08} & 6.36 \\
NRKNet (`24)~\citep{quan2023neumann} &
\textcolor{darkgreen}{27.60} & \textcolor{darkgreen}{28.37} & 27.21 & 26.16 & 26.33 &
0.90 & \textcolor{darkgreen}{0.89} & 0.86 & 0.86 & 0.86 &
0.06 & 0.07 & 0.10 & 0.14 & 0.15 &
\textcolor{darkgreen}{27.13} & 0.87 & 0.10 & \textcolor{darkgreen}{0.35} \\
ViTDeblur (`24)~\citep{liang2024vitdeblur} &
27.22 & 27.87 & \textcolor{darkgreen}{27.27} & \textcolor{darkgreen}{26.37} & \textcolor{darkgreen}{26.70} &
0.90 & \textcolor{darkgreen}{0.89} & \textcolor{darkgreen}{0.87} & 0.87 & \textcolor{darkgreen}{0.87} &
0.06 & \textcolor{darkgreen}{0.06} & \textcolor{darkgreen}{0.08} & \textcolor{darkgreen}{0.10} & 0.11 &
27.09 & \textcolor{darkgreen}{0.88} & \textcolor{darkgreen}{0.08} & 12.91 \\
\midrule

%% ────────────────────────── MOTION DEBLURRING ──────────────────────────
\multicolumn{20}{l}{\textbf{Motion Deblurring Methods} (stress-tested on \textit{Scene 1})} \\
\midrule
EVSSM (`25)~\citep{kong2025efficient} &
22.64 & 23.61 & 18.76 & 17.03 & 16.55 &
0.81 & 0.85 & 0.76 & 0.72 & 0.70 &
0.10 & 0.08 & 0.13 & 0.19 & 0.24 &
19.34 & 0.76 & 0.16 & -- \\
FFTformer (`23)~\citep{kong2023efficient} &
23.44 & 24.78 & 19.58 & 17.44 & 16.97 &
0.82 & 0.85 & 0.77 & 0.73 & 0.71 &
0.12 & 0.11 & 0.17 & 0.23 & 0.28 &
20.03 & 0.77 & 0.19 & -- \\
\bottomrule
\end{tabular}}
\label{tab:deblur_results_full}
\vspace{-10pt}
\end{table}

We employ MODEST for evaluating state of the art methods in two practical applications: shallow DoF rendering (Section~\ref{subsec:shallow_dof}) and defocus deblurring (Section~\ref{subsec:deblur}). All pre-trained models are tested in zero-shot manner on 4 scenes from MODEST after performing a detailed search for optimal tunable parameter settings for each method. Parameter tuning and sensitivity analysis are in Sections~\ref{subsec:tuning} and \ref{subsec:sensitivity} respectively. Additional experiments, qualitative results and visuals with other scenes are shown in the supplementary material.\vspace{-6pt}  %We also show results %Sections below use Scene 1 as the primary test case for quantitative evaluation due to its optical complexity. 

\subsection{Shallow DoF Rendering Models}\label{subsec:shallow_dof}

\begin{figure}[htbp]  % * makes it span both columns
    \centering
    \setlength{\abovecaptionskip}{1pt}  % reduce space above caption
    \setlength{\belowcaptionskip}{1pt} % optional: reduce space below caption    
    \includegraphics[width=0.99\textwidth]{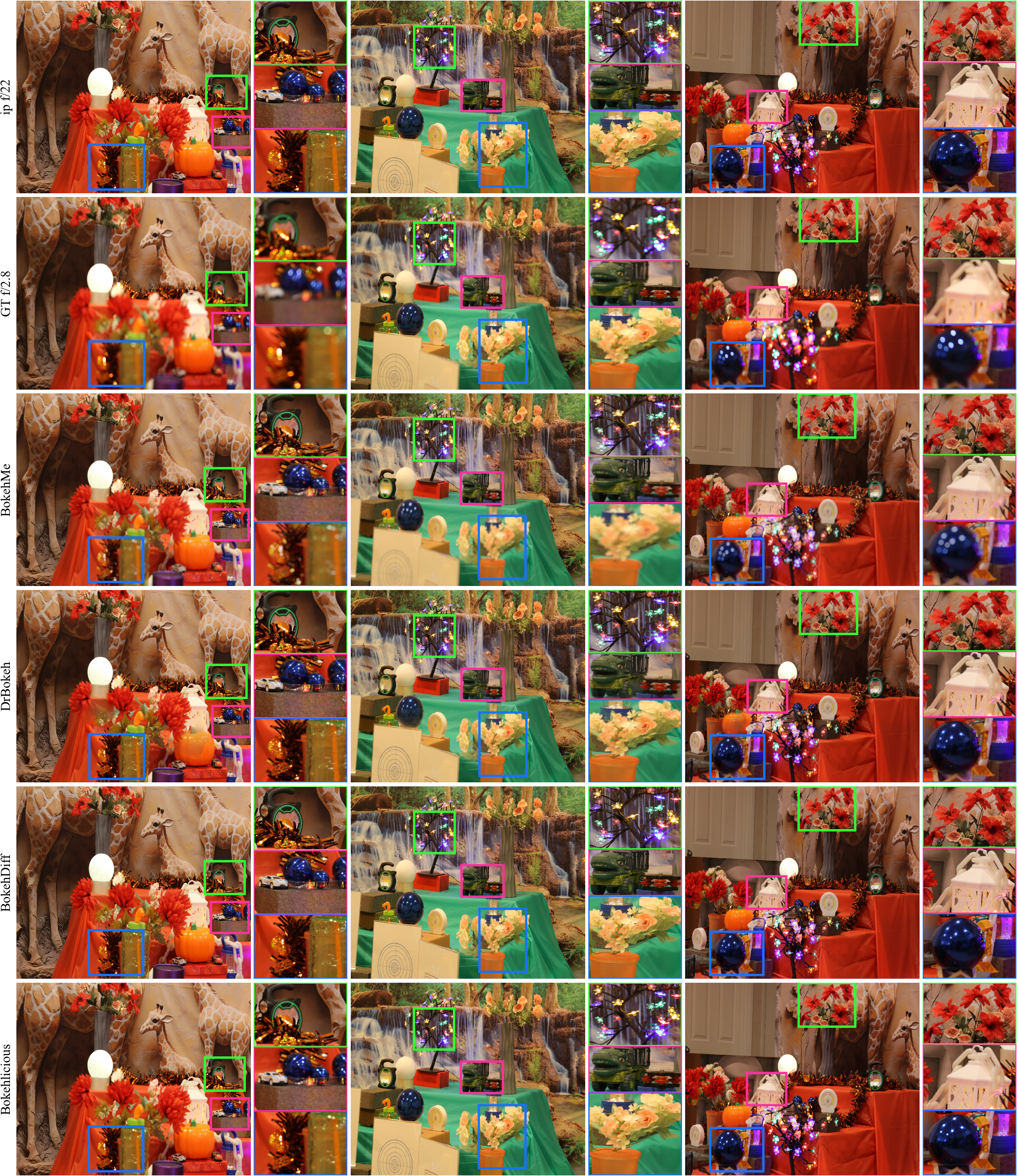}   
    % scale to text width    
    \caption{Four SOTA shallow depth of field (DoF) rendering methods for three different MODEST images for focal length of 70mm. Sharp input is f/22.0 aperture, output target is f/2.8 aperture.}
    \label{fig:dof}        
\end{figure}

To evaluate the generalization capability of shallow depth of field (DoF) rendering methods on MODEST, we benchmarked four representative state-of-the-art models: BokehMe~\citep{peng2022bokehme}, Dr.Bokeh~\citep{sheng2024dr}, BokehDiff~\citep{zhu2025bokehdiff} and Bokehlicious~\citep{seizinger2025bokehlicious}. Each model is evaluated on four scenes at five focal lengths covering narrow to wide field of view (FoV). 20 images per focal length $\times$ 5 focal lengths $\times$ four scenes, 400 images total are used for benchmarking the four DoF methods. 

For each model at each focal length, we run a multi-parameter grid-search across all tunable parameters, each time running inference on all images of a scene and gathering PSNR, LPIPS, SSIM metrics. This ensures optimal parameters are identified for benchmarking of these methods at their best performance at each focal configuration. E.g., for BokehMe, blur-strength K$\in$[40,60], normalized focus plane disparity $\in$[0.15,0.55], defocus scale $\in$[10,30] were grid-traversed at each of 5 focal lengths at all images of a scene, taking about 63 hours to find the best performing configuration. For Dr.Bokeh, grid traversal for blur-strength K$\in$[15,35] and normalized focus plane disparity $\in$[0.2,0.6] took about 50hours to find the best performing configuration. With this extensive tuning protocol, our focus has been to evaluate the best performance of each method while covering the full range of field of view of the camera, the widest at 70mm and narrowest at 28mm.

\begin{figure}[htbp]
\centering
 \setlength{\abovecaptionskip}{1pt}  % reduce space above caption
    \setlength{\belowcaptionskip}{1pt} % optional: reduce space below caption  
\begin{subfigure}{0.33\textwidth}
    \centering
    \includegraphics[width=\linewidth]{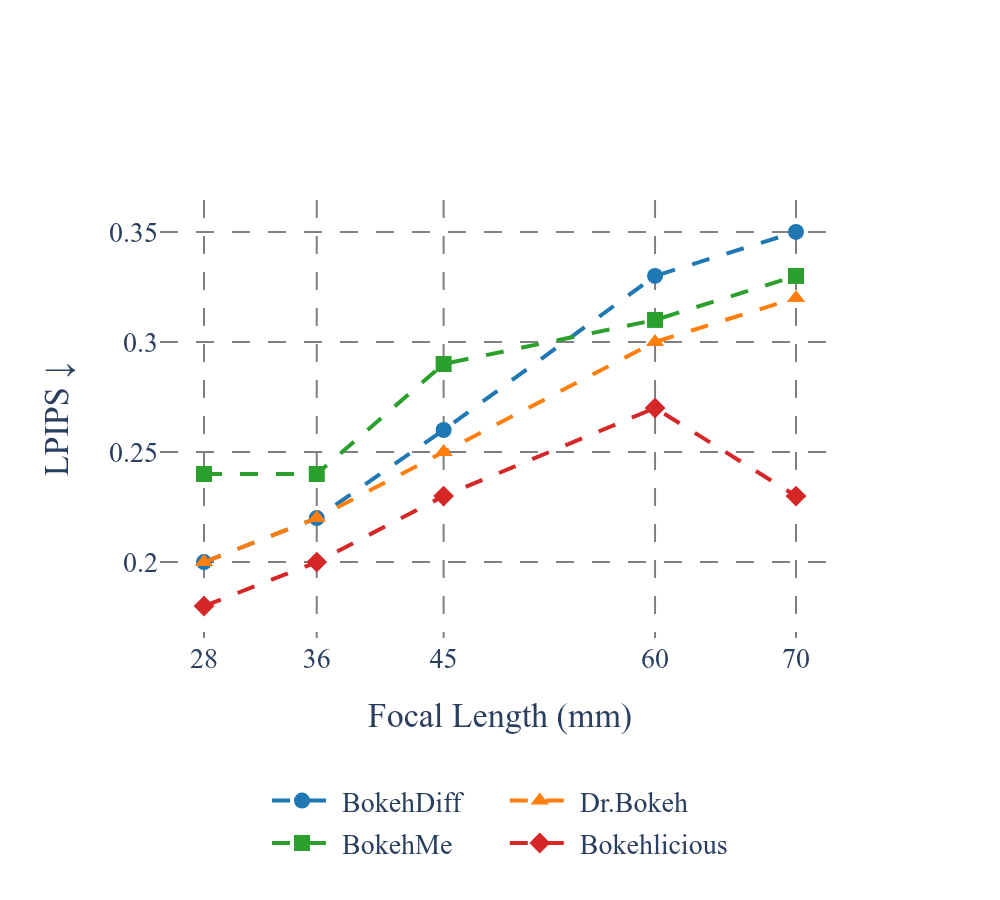}
    \caption{Depth of field methods vs focal length}
\end{subfigure}\hfill
\begin{subfigure}{0.33\textwidth}
    \centering
    \includegraphics[width=\linewidth]{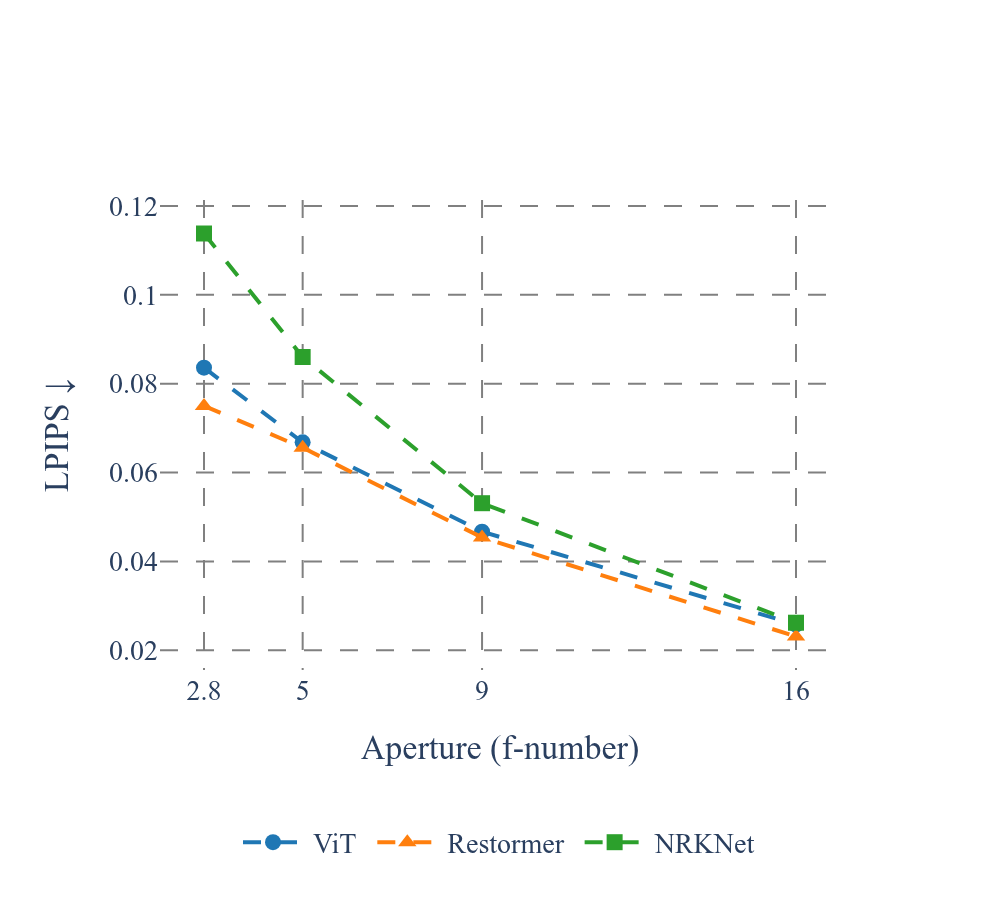}
    \caption{Deblurring methods vs aperture}
\end{subfigure}\hfill
\begin{subfigure}{0.33\textwidth}
    \centering
    \includegraphics[width=\linewidth]{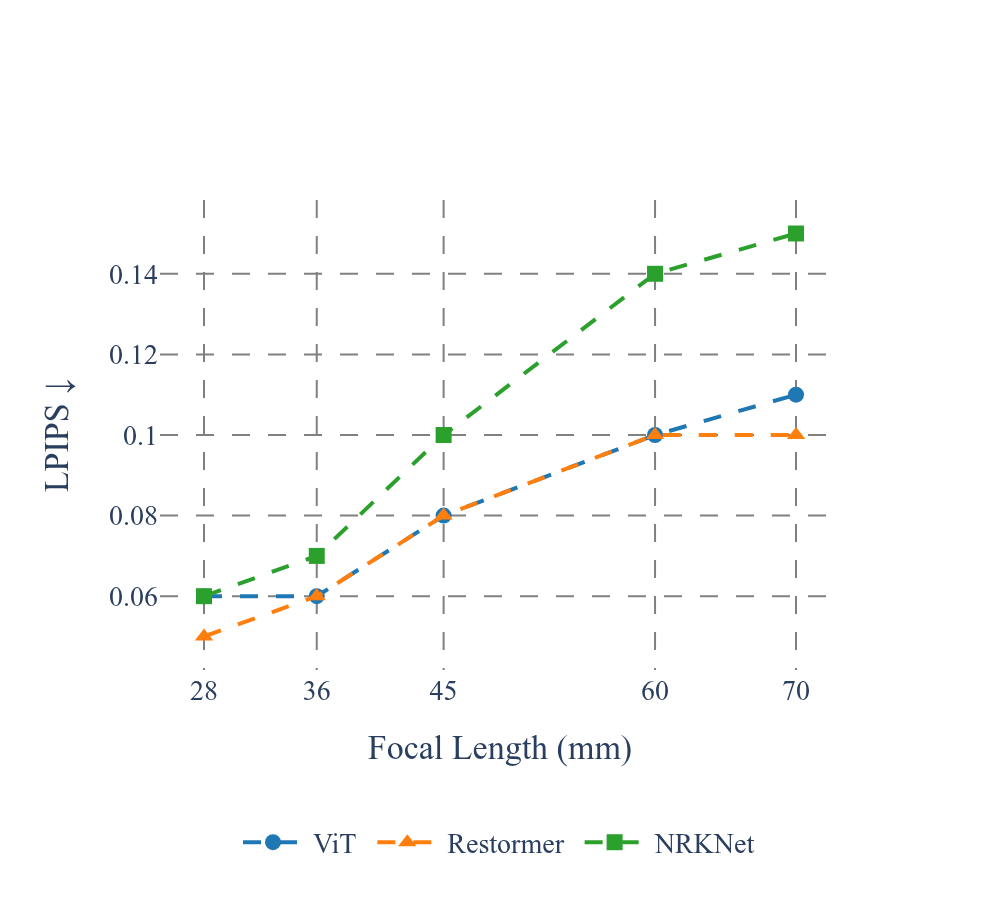}
    \caption{Deblurring methods vs focal length}
\end{subfigure}

\caption{Ablation study of 5 depth of field and 3 defocus deblurring methods across imaging parameters. Lower LPIPS is desired.}
\label{fig:lpips_comparison_all}
\vspace{-10pt}
\end{figure}

At each focal length, the DoF evaluation task is to synthesize shallow DoF images at f/2.8 using the corresponding sharp all-in-focus input image captured at f/22.0. These apertures bound the maximum defocus range and give the clearest, most diagnostic comparison among methods at each focal length. As described in Sections~\ref{subsec:tuning} and \ref{subsec:sensitivity}, while current DoF parameters are very time consuming to tune, as better architectures emerge, MODEST enables benchmarking them at any input-output aperture combination amongst f/22.0, f/16.0, f/9.0, f/5.0, and f/2.8. For example, we use MODEST data to evaluate defocus deblur methods on multiple apertures as shown in Fig.~\ref{fig:lpips_comparison_all}(b).   

The GT blurry target and all model outputs are aligned with the sharp input to calculate metrics without pixel shift. The Tab.~\ref{tab:dof_results} shows that in four-scene average, Bokehlicious performs the best in all 3 metrics. With a clear margin, Dr.Bokeh is the second best in all 3 metrics. BokehDiff performs the worst in PSNR and SSIM and is never the optimal method in any of the four scenes. Additionally we gathered human rankings from 10 reviewers for the 4 DoF models on 53 different images each, across five focal lengths, yielding 530 total rankings, shown in Mean Opinion Rank (MOR) column besides the inference time on A100 GPU per image in Tab.~\ref{tab:dof_mor}. Lower MOR indicates better perceptual quality. BokehMe and Bokehlicious outputs were favored the most, while BokehDiff was the least favorite. As noted in ~\citep{ignatov2020rendering}, visual perception and human-rankings can differ from numeric metrics like PSNR, SSIM, LPIPS, as they capture perceptual quality rather than pixel-level fidelity. This is particularly relevant for DoF rendering, where rendering realism depends on blur smoothness, bokeh characteristics, and edge transitions, which are not fully captured by numerical metrics.

  MODEST enables structured analysis of DoF across focal lengths. Fig.~\ref{fig:lpips_comparison_all}(a) shows that the performance degrades as focal length increases from \texttt{fl28} to \texttt{fl70}, even with fixed aperture, indicating that the current models fail to scale the Circle of Confusion (CoC) under changing magnification and field of view. Note that each result was obtained with parameters tuned at that focal length, i.e., we are evaluating the best possible performance per method. Visual results in Fig.~\ref{fig:dof} highlight artifacts such as exaggerated blur, incorrect focal plane placement, and inconsistent depth ordering, including cases where background regions are sharpened while foreground is blurred. These limitations may arise from training on small-resolution, limited optics datasets such as EBB!~\citep{ignatov2020rendering} and BLB~\citep{bokehme_blb}, which lack explicit optical structure, leading models to learn appearance–blur correlations rather than optical principles. MODEST provides stereo imagery with inferrable focus planes, multiple focal lengths, and systematically varied apertures, making the relationship between depth, aperture, and blur explicit and controlled. The use of two synchronized full-frame Canon EOS 6D cameras further preserves authentic lens characteristics such as realistic bokeh shapes and depth dependent blur.

Fig.~\ref{fig:dof} and supplementary figures show that almost all methods produce uniform or Gaussian density blur effect. Dr.Bokeh, BokehDiff, BokehMe models do not take focal length or aperture as explicit inputs but instead need tuning for non-intuitive parameters such as K (blur radius), blur strength, normalized focus plane disparity. Even at the optimally tuned parameters, these models blur regions spanning wide range of depth or keep majority of image sharp. Bokehlicious does not always match the extent of blur to the GT, but generally performs better than the other methods at pixel-level details in sharp regions which can be attributed to being trained at higher resolution. 

Dr.Bokeh combines CNN-based foreground and background segmentation with kernel-based rendering using RGB-D and camera parameters, yet fails to identify the focal plane, resulting in inconsistent blur, likely due to limited modeling of CoC, lens encoding, and sensor characteristics. BokehMe employs CNN based transformer features with point spread function (PSF) based rendering, but often produces uniform blur in wrong regions, indicating weak focal plane estimation and poor modeling of texture-dependent defocus. Bokehlicious introduces aperture aware attention and lens encoding, improving texture based blur, but the absence of depth input limits its ability to resolve the focal plane accurately. BokehDiff integrates depth, lens parameters, and CoC within a diffusion framework, achieving more reliable focal plane localization and better geometric consistency. However, its reliance on synthetic training data constrains realistic blur. Overall, BokehDiff captures scene structure more effectively, while Bokehlicious yields more consistent texture aware blur, reflecting a trade-off between geometric accuracy and visual fidelity.

Fig.~\ref{fig:lpips_comparison_all}(a) shows the LPIPS analysis of different DoF methods across focal lengths. As focal length increases, LPIPS rises for all methods, indicating that realistic depth of field rendering becomes more challenging at narrower FoVs, particularly in preserving sharp depth transitions and consistent blur. Among the methods, Bokehlicious consistently achieves the lowest LPIPS. Supplementary material demonstrates failure of these SOTA DoF models to render cat-eye or non-uniform-intensity bokeh shapes, and high variance in outputs for near-identical multi-capture inputs of MODEST.

An important distinction of MODEST is availability of stereo data absent in all prior DoF datasets. Using the calibrated stereo disparity information from left and right views of the same image and FoundationStereo~\citep{foundation}, a SOTA stereo depth model, we obtained stereo depth map. For the first time in DoF literature, we stress tested these DoF methods to run with stereo depth maps instead of monocular depth maps for all five focal lengths for Scene 1, without any additional parameter tuning. Despite all DoF methods being trained on monocular images and depth maps, and no additional tuning of the configurable parameters, we observed that Scene 1 stereo results in Tab.~\ref{tab:stereo_dof_results} landed better than Scene 1 monocular results in PSNR and very close in SSIM and LPIPS. Most notably, Dr.Bokeh saw the PSNR gain of 3.5dB with stereo depth information. Because the models were never trained on stereo-derived depth, this improvement can be attributed to the higher geometric fidelity of stereo depth relative to monocular estimates, particularly at close range and around depth discontinuities. This significant result confirms that the stereo viewpoints in MODEST carry information that off-the-shelf monocular depth cannot fully recover and establishes strong motivation for stereo-aware and stereo-native datasets like MODEST for modern DoF applications in AR glasses, VR headsets, multi-lens smartphone cameras, drones, autonomous robots and more.

We compared single-image inference time as a mean of 3 inferences after warm-up on an A100-40GB GPU. Bokehlicious is the fastest at 17.8s, Dr.Bokeh is the slowest at 120.4s. For BokehMe and BokehDiff, the total time is the sum of depth estimation and bokeh rendering times in order, as shown in Tab.~\ref{tab:dof_mor}.\vspace{-6pt}

\subsection{Deblurring Models}\label{subsec:deblur}

\begin{figure}[htbp]
    \centering
     \setlength{\abovecaptionskip}{1pt}  % reduce space above caption
    \setlength{\belowcaptionskip}{1pt} % optional: reduce space below caption  
    \includegraphics[width=0.98\textwidth]{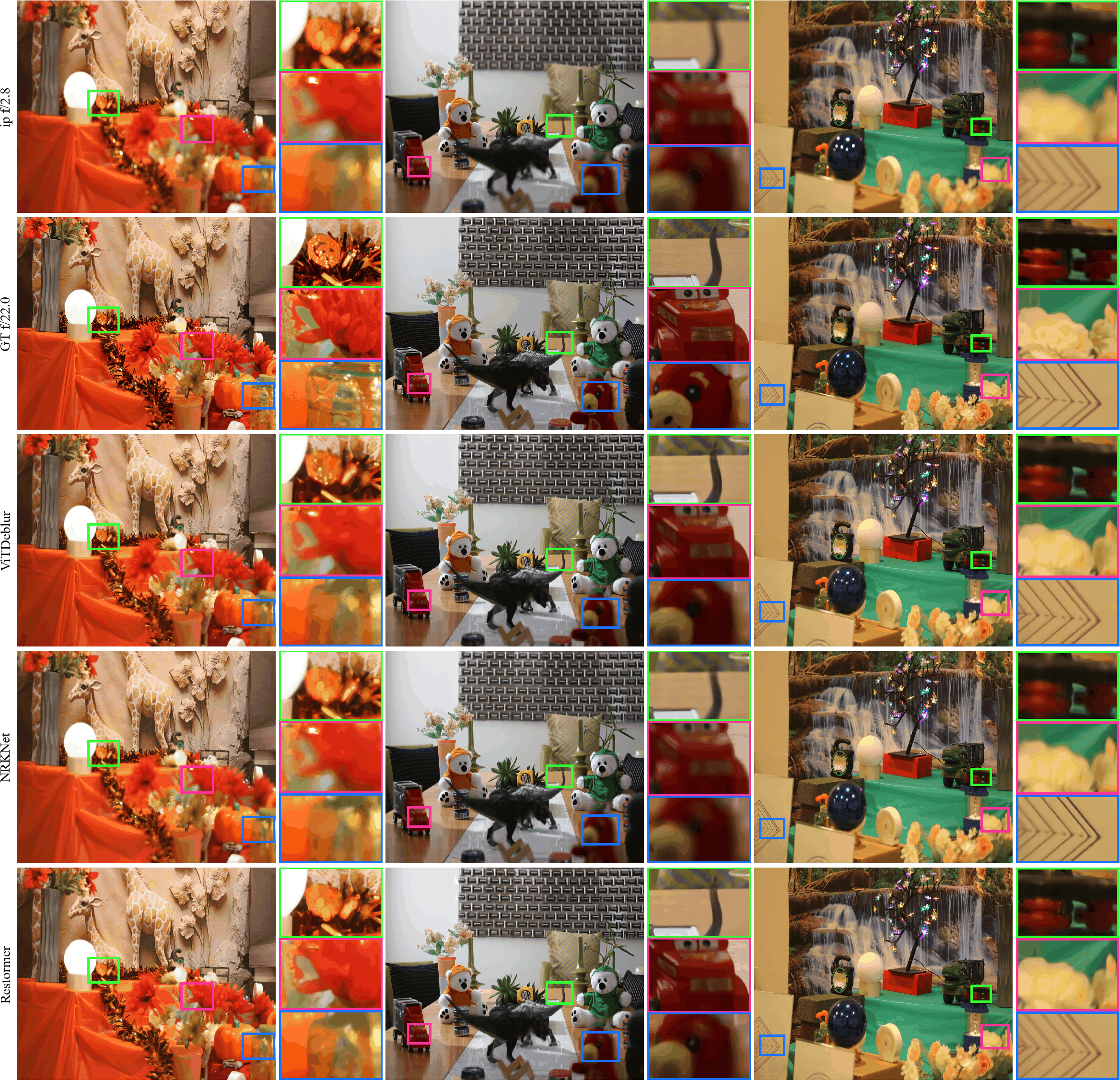}
    \caption{
    Qualitative comparison of three defocus deblurring models across three different MODEST scenes for a focal length of 70mm. The input images are captured at a wide aperture ($f/2.8$), and the ground-truth images correspond to sharp all-in-focus captures at f/22.0.
    }
    \label{fig:deblur_70mm}
    \vspace{-10pt}
\end{figure}

To mitigate aperture-induced defocus blur, we benchmark three defocus deblurring models: Restormer~\citep{zamir2022restormer}, ViTDeblur~\citep{liang2024vitdeblur}, and NRKNet~\citep{quan2023neumann}. These methods are designed to recover sharp images from spatially varying defocus blur and therefore represent strong baselines for this task. For completeness, we stress-test two motion-deblurring architectures, EVSSM~\citep{kong2025efficient} and FFTformer~\citep{kong2023efficient}. Originally trained on motion-blur datasets, these two models are only assessed for how motion-trained networks generalize to optical defocus.

Quantitative evaluation is performed on 4 scenes, comprising 20 images for each of five focal lengths (\texttt{fl28}-\texttt{fl70} mm), resulting in a total of 400 images. The blurred input is captured at the widest aperture (f/2.8), while the corresponding all-in-focus GT reference image is captured at f/22.0. Input and output images are aligned to avoid any pixel shifts. MODEST provides real, depth-dependent defocus blur generated by the camera optics rather than synthetic blur kernels. Tab.~\ref{tab:deblur_results_full} shows 4-scene performance for the widest aperture input (f/2.8), also shown as trend lines in  Fig.~\ref{fig:lpips_comparison_all}(c). NRKNet achieves high PSNR, usually at lower focal lengths (wide FoV). At higher focal lengths (narrow FoV), NRKNet shows sharp degradation in LPIPS. As focal length increases, LPIPS degrades for all methods, reflecting increased difficulty in recovering sharp details under stronger defocus. Restormer and ViTDeblur remain close in almost all configurations. NRKNet is significantly faster than the other two methods and suits real-time deployment better. Fig.~\ref{fig:lpips_comparison_all}(b) shows that the performance improves as the aperture narrows and input image gets sharper. This is expected as the methods need to recover less of the blurred information from the relatively sharper input.

Deblurring methods are the exact opposite of DoF, where the input is a defocused blurry image and the expected output is a sharp image. This is quite challenging as it involves restoring pixels that are lost due to defocus. When compared visually, Restormer does a good job of removing blur in simple scenes but fails in more complex ones. We observed many occurrences of Restormer struggling when the scene is at a closer distance with higher blur. 

NRKNet uses a Neumann architecture ~\citep{gilton2019neumannnetworksinverseproblems} for feature extraction along with physics-based lens encoding, but it still does not deblur in many regions. This could indicate that the architecture is not sufficient to extract detailed features required for effective deblurring. This is where ViT-based methods like VitBlur come in, using vision transformers that divide the image into patches to capture fine details. However, when zooming into the output, patch boundaries become visible as small overlapping boxes as visible in supplementary figures. While it performs comparatively better in deblurring, using higher resolution training data like MODEST and improved boundary-stitching mechanisms could further enhance the final output. An interesting observation across all deblurring methods is that they tend to fail when objects are very close to each other, often struggling to separate them and leaving residual blur. Additionally, they show noticeable degradation when restoring text and handling light reflections.

Qualitative results in Fig.~\ref{fig:deblur_70mm} further illustrate the difficulty of the task. While most methods produce visually plausible outputs at a coarse scale, magnified regions reveal persistent artifacts around fine structures, reflective surfaces, transparent objects, and small light sources. These errors arise from the spatially varying and depth-dependent nature of defocus blur, which differs fundamentally from the synthetic blur data or artifical blur kernels typically used during training. 

% \input{text/4_calibration} % (Commented out in your original code)
%\newpage
% \section{tuning}
\subsection{DoF Methods: Tuning}\label{subsec:tuning}

We perform grid search for optimal hyperparameters for all DoF methods at each focal length. Boekhlicious has only one tunable parameter $\mathcal{F}$. The other three methods have two or more parameters, needing two-stage grid search: first we optimize for the scene-related parameter, such as focus plane disparity, at a fixed focal length. Then, we optimize for camera-related parameter such as blur strength per focal length. At each cell in the grid traversal, we run inference on all images from one scene at that focal length for input image aperture of $F22$ and output target aperture of $F2.8$ and compute PSNR, SSIM, LPIPS metrics. Such extensive tuning experiments ensure that the best parameters are found and used in benchmarking in Tab. \ref{tab:dof_results}. In case of slight ambiguity among PSNR, SSIM and LPIPS, LPIPS is preferred. Tab. \ref{tab:optimal_hyperparameters} records optimal parameters for each method.

\begin{figure}[htbp]
\centering
\setlength{\abovecaptionskip}{1pt} % reduce space above caption
\setlength{\belowcaptionskip}{1pt} % optional: reduce space below caption  

\begin{subfigure}{0.95\textwidth}
    \centering
    \includegraphics[width=\linewidth]{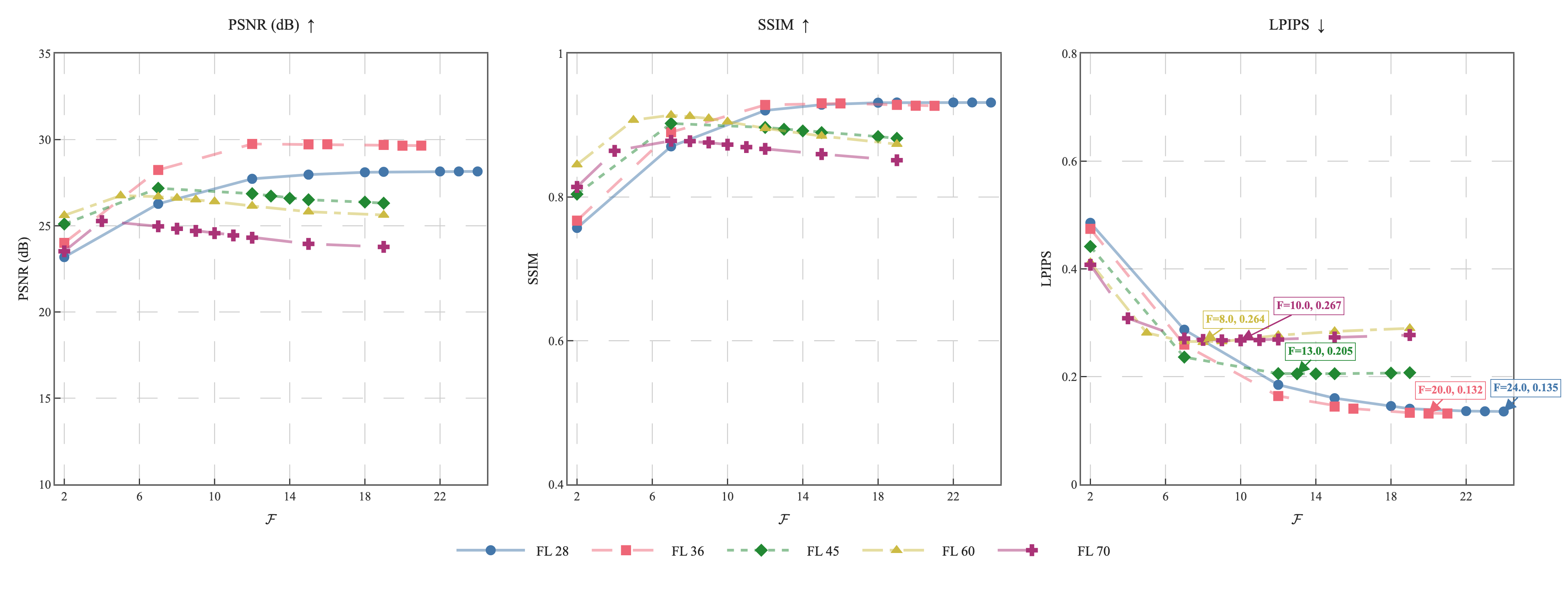}
    \vspace{-20pt}
    \caption{Bokehlicious: Best $\mathcal{F}$-parameter search}
\end{subfigure}\hfill

\begin{subfigure}{\textwidth}
    \centering
    \includegraphics[width=\linewidth]{imgs_fullsz/bokehme_stage_1.png}
    \vspace{-20pt}
    \caption{BokehMe stage-1: Best defocus scale and normalized focus plane disparity search}
\end{subfigure}\hfill

\begin{subfigure}{0.95\textwidth}
    \centering
    \includegraphics[width=\linewidth]{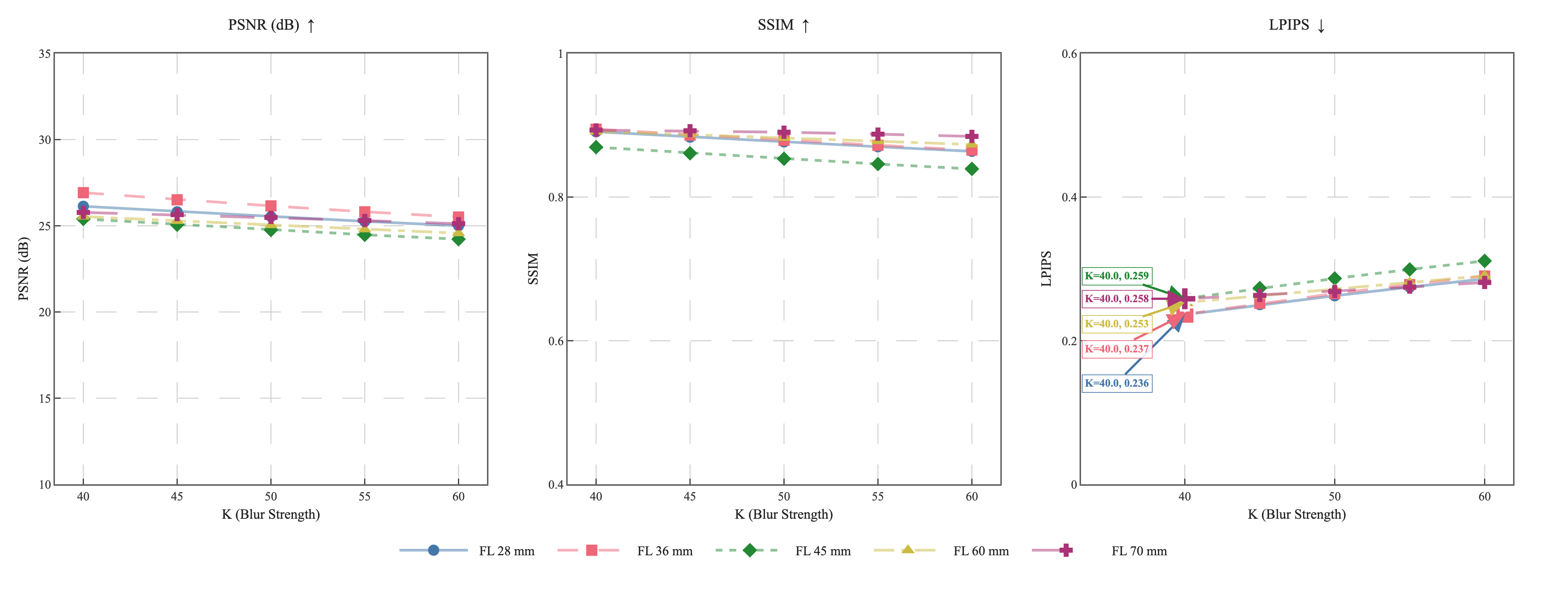}
    \vspace{-20pt}
    \caption{BokehMe stage-2: Best blur strength search}
\end{subfigure}\hfill

\caption{
Hyperparameter search for the Bokehlicious and BokehMe DoF methods. Each plot reports the PSNR, SSIM and LPIPS obtained for different hyperparameter configurations.
}

\label{fig:tuning_bokehlicious_bokehme}
\end{figure}

Fig.~\ref{fig:tuning_bokehlicious_bokehme} (a) shows optimal $\mathcal{F}$-parameter for each focal length. Note that input aperture for all focal lengths is $F22$. However, the optimal $\mathcal{F}$-parameter for each focal length changes: $\mathcal{F}=24$ for fl 28mm and $\mathcal{F}=10$ for fl 70mm. This trend is nonlinear and nonintuitive, i.e., the tunable parameter $\mathcal{F}$ has no relation with actual input image aperture F. 

Fig.~\ref{fig:tuning_bokehlicious_bokehme} (b) shows stage-1 hyperparameter search of the BokehMe method for normalized focus plane disparity (dispfocus) and defocus-strength (DFS). With the optimal values of dispfocus=0.15 and dfs=20, Fig.~\ref{fig:tuning_bokehlicious_bokehme} (c) shows stage-2 hyperparameter search of the BokehMe method for blur strength (K) at each of five focal lengths. Contrary to the optical intuition that the blur radius should change for different FoVs, the optimal K parameter for the BokehMe method does not changea across focal lengths. 

\begin{figure}[htbp]
\centering
\setlength{\abovecaptionskip}{0.5pt}  % reduce space above caption
\setlength{\belowcaptionskip}{0.5pt} % optional: reduce space below caption  

\begin{subfigure}{\textwidth}
    \centering
    \includegraphics[width=\linewidth]{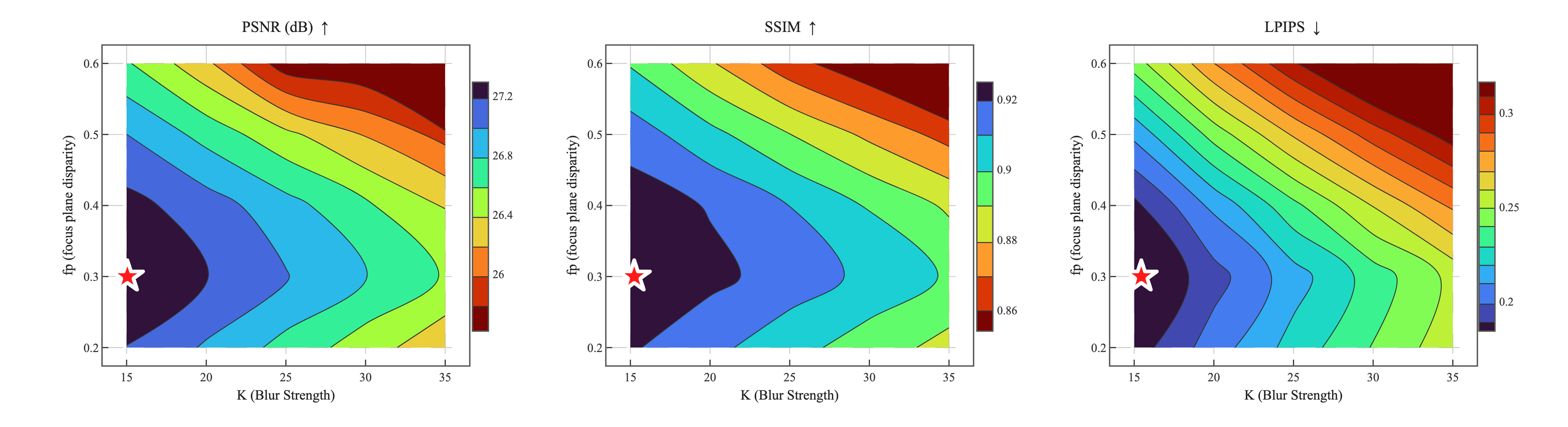}
    \vspace{-25pt}
    \caption{Dr.Bokeh stage-1: Best normalized focus plane disparity search}
\end{subfigure}\hfill

\begin{subfigure}{0.9\textwidth}
    \centering
    \includegraphics[width=\linewidth]{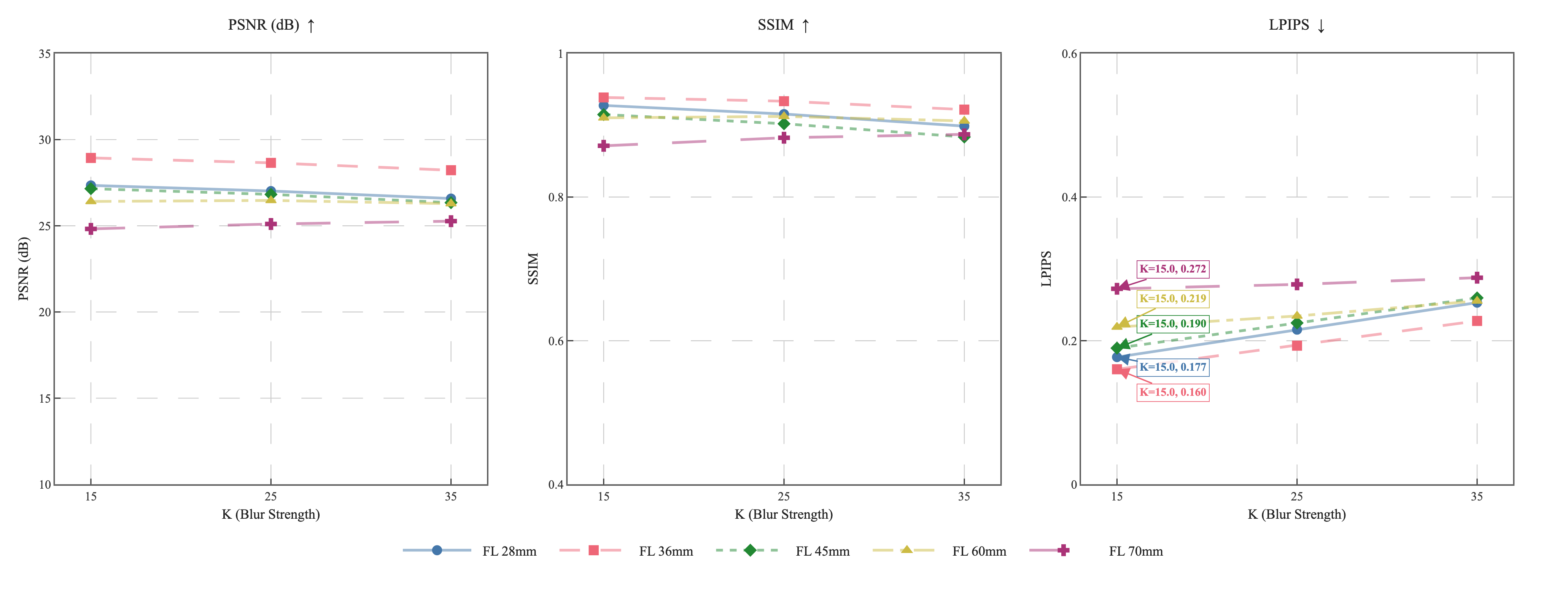}
    \vspace{-25pt}
    \caption{Dr.Bokeh stage-2: Best blur strength (K) search}
\end{subfigure}\hfill

\begin{subfigure}{\textwidth}
    \centering
    \includegraphics[width=\linewidth]{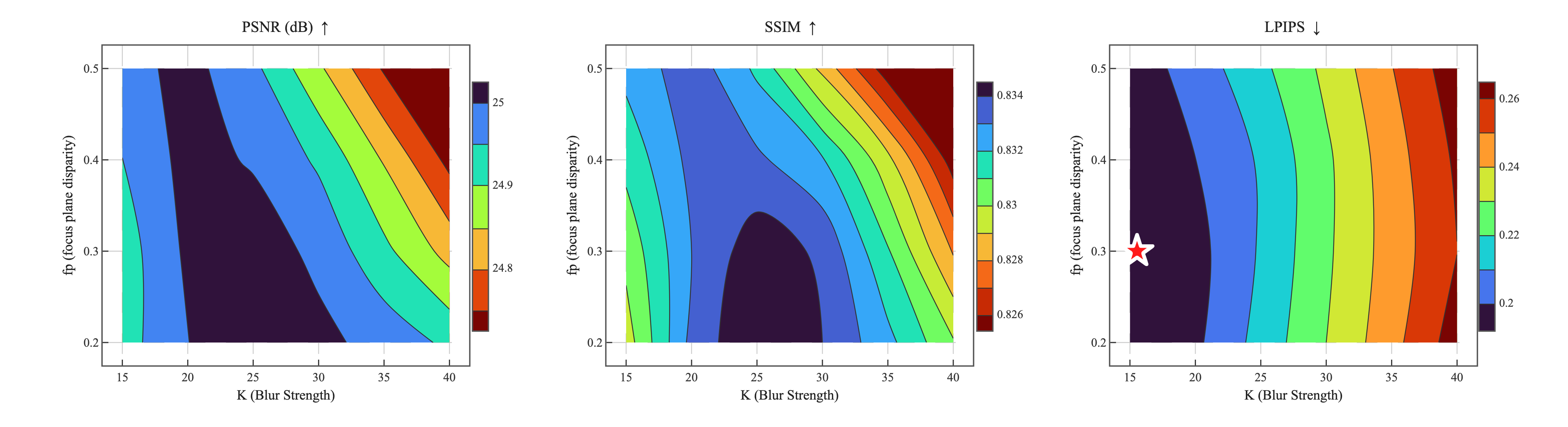}
    \vspace{-25pt}
    \caption{Bokehdiff stage-1: Best normalized focus plane disparity search}
\end{subfigure}\hfill

\begin{subfigure}{0.9\textwidth}
    \centering
    \includegraphics[width=\linewidth]{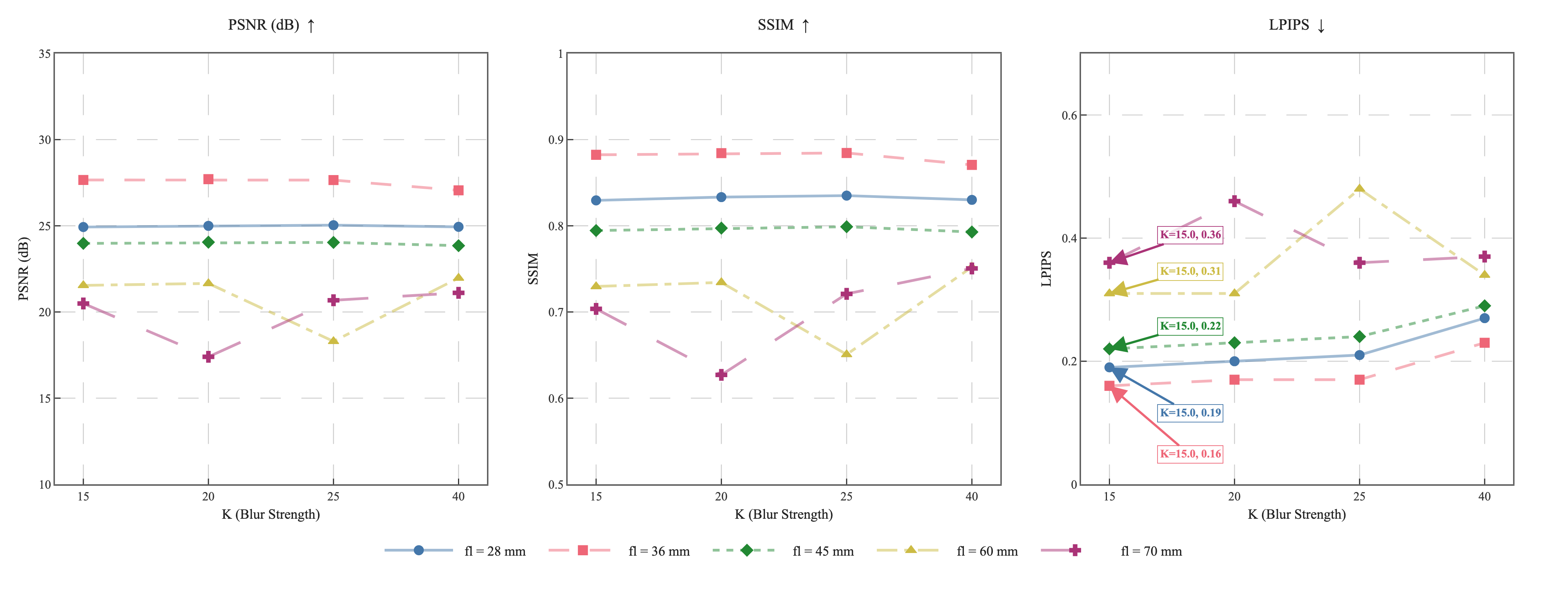}
    \vspace{-25pt}
    \caption{Bokehdiff stage-2: Best blur strength search}
\end{subfigure}\hfill

\caption{Hyperparameter search for the Dr.Bokeh and Bokehdiff DoF methods. Each plot reports the PSNR, SSIM and LPIPS obtained for different hyperparameter configurations.}
\label{fig:tuning_drbokeh_bokehdiff}
\end{figure}

Fig.~\ref{fig:tuning_drbokeh_bokehdiff} (a),(c) show stage-1 hyperparameter search of the Dr.Bokeh and BokehDiff methods for normalized focus plane disparity. With the optimal value of fp=0.3, Fig.~\ref{fig:tuning_drbokeh_bokehdiff} (b),(d) show stage-2 hyperparameter search of the Dr.Bokeh and BokehDiff methods for blur strength (K) at each of five focal lengths. Both methods share the trend that the optimal K parameter does not change proportionally to the change in FoV with focal length, contrary to how optical depth of field phenomenon works.

With these extensive tuning experiments, evidence suggests that the SOTA DoF methods have non-intuitive parameters whose optimal values do not tune proportionally to actual image attributes like focal length. Optimal blur strength for three methods BokehMe, Dr.Bokeh and BokehDiff show insensitivity to focal length. 
Bokehlicious $\mathcal{F}$-parameter has no relation with actual input image aperture F and it shows nonlinear trend with input focal length at a constant aperture.

% Table
\begin{table}[htbp]
\centering
\footnotesize
\setlength{\tabcolsep}{3pt}
\caption{Optimal inference hyperparameters used for evaluating each depth of field  model on the MODEST dataset\vspace{-1ex}}
\label{tab:optimal_hyperparameters}
\begin{tabular}{l|ccc}
\hline
\textbf{Model} & \textbf{Para. 1} & \textbf{Para. 2} & \textbf{Para. 3} \\ \hline
%\textbf{Bokehlicious} & \parbox[c]{3.2cm}{\vspace{1mm} 
%$F=24$ (FL 28), $F=20$ (FL 36)\\
%$F=13$ (FL 45), $F=8$ (FL 60)\\
%$F=10$ (FL 70)\vspace{1mm}} & $-$ & $-$ \\ \hline
\textbf{Bokehlicious} & \begin{tabular}[c]{@{}l@{}} $F$: 24 (FL28), 20 (FL36), 13 (FL45), \\ 8 (FL60), 10 (FL70) \end{tabular} & $-$ & $-$ \\ \hline
\textbf{BokehMe} & $dispfocus = 0.15$ & $K = 40 $ & $dfs = 20$ \\ \hline
\textbf{Dr.Bokeh} & $fp = 0.3$ & $K = 15$ & $-$ \\ \hline
\textbf{Bokehdiff} & $fp = 0.3$ & $K = 15$ & $-$ \\ \hline
\end{tabular}\vspace{-1ex}
\end{table}

% FL 28 -> F=24, FL 36 -> F=20, FL 45 -> F=13, FL 60 -> F= 8, FL 70 -> F=10.

%\newpage
\subsection{DoF Methods: Sensitivity Analysis}\label{subsec:sensitivity}

\begin{figure}[htbp]
\centering
\setlength{\abovecaptionskip}{1pt}  % reduce space above caption
\setlength{\belowcaptionskip}{1pt} % optional: reduce space below caption  

\begin{subfigure}{0.95\textwidth}
    \centering
    \includegraphics[width=\linewidth]{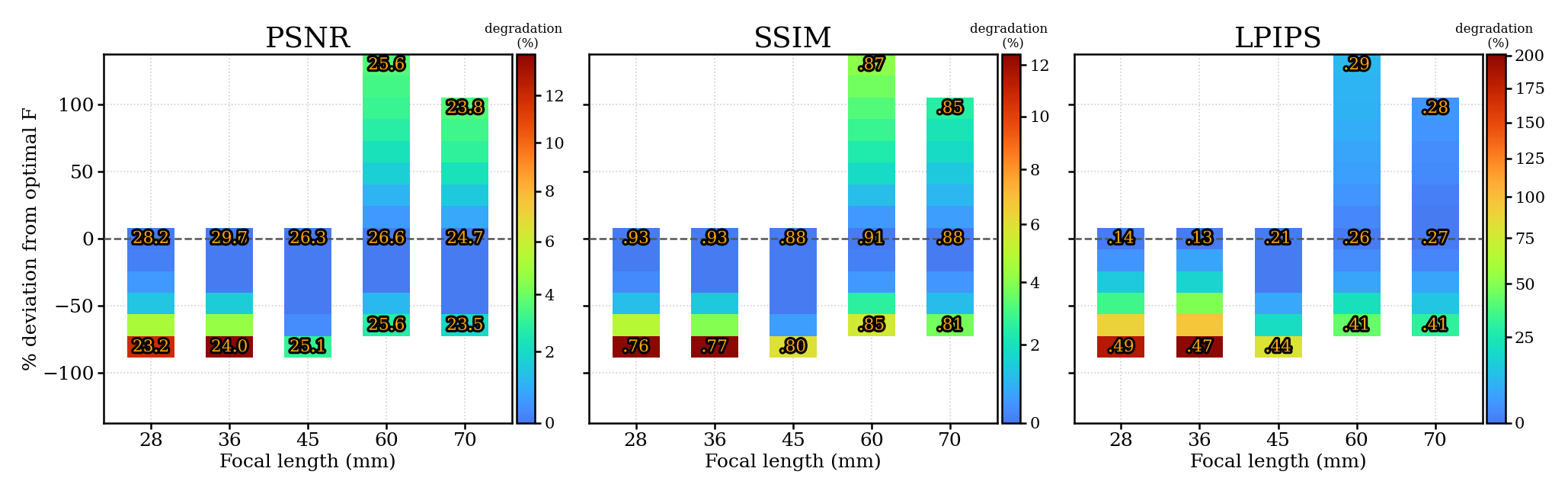}
     \vspace{-20pt}
    \caption{Bokehlicious sensitivity plot.}
\end{subfigure}\hfill

\begin{subfigure}{0.95\textwidth}
    \centering
    \includegraphics[width=\linewidth]{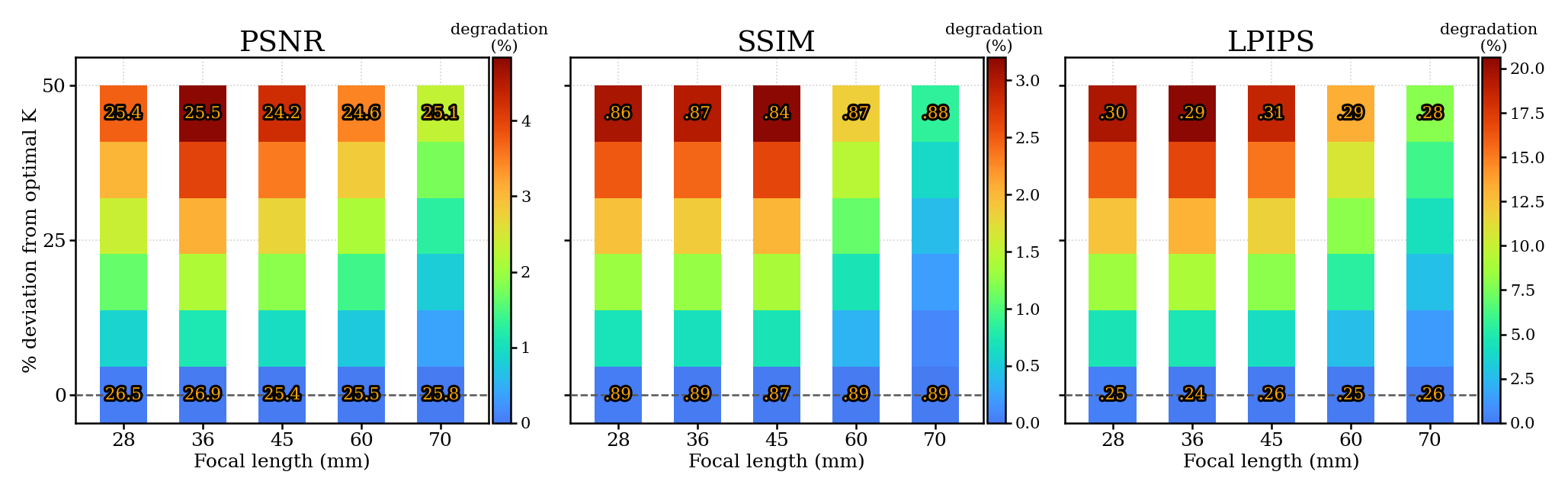}
     \vspace{-20pt}
    \caption{BokehMe sensitivity plot.}
\end{subfigure}\hfill

\begin{subfigure}{0.95\textwidth}
    \centering
    \includegraphics[width=\linewidth]{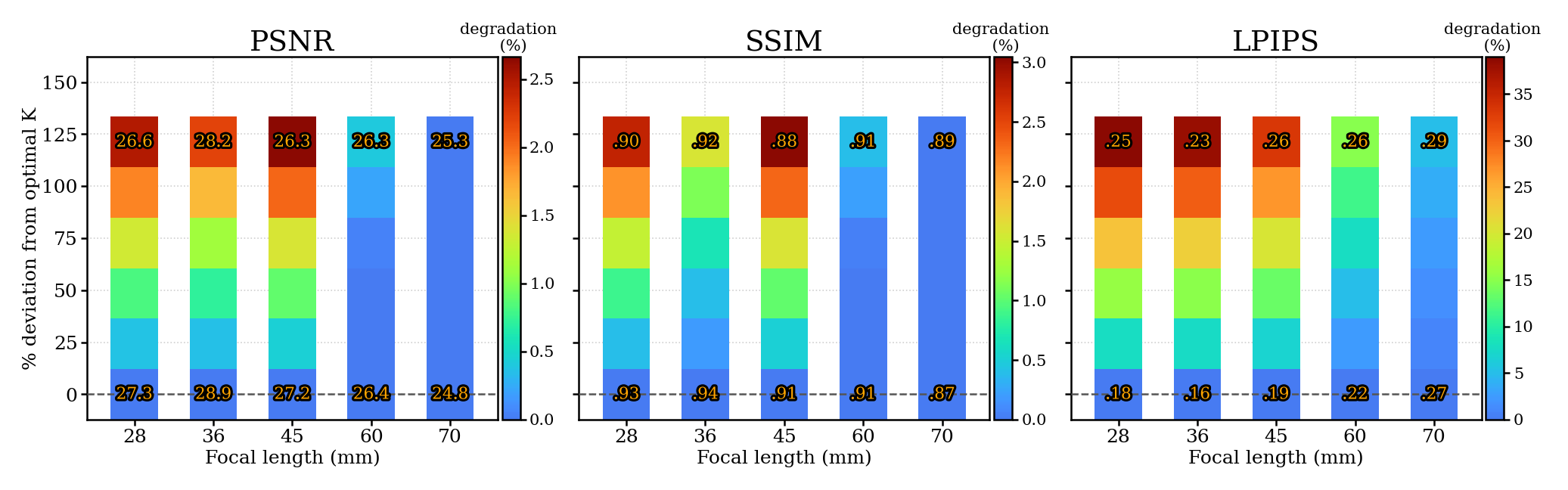}
     \vspace{-20pt}
    \caption{Dr.Bokeh sensitivity plot.}
\end{subfigure}\hfill

\begin{subfigure}{0.95\textwidth}
    \centering
    \includegraphics[width=\linewidth]{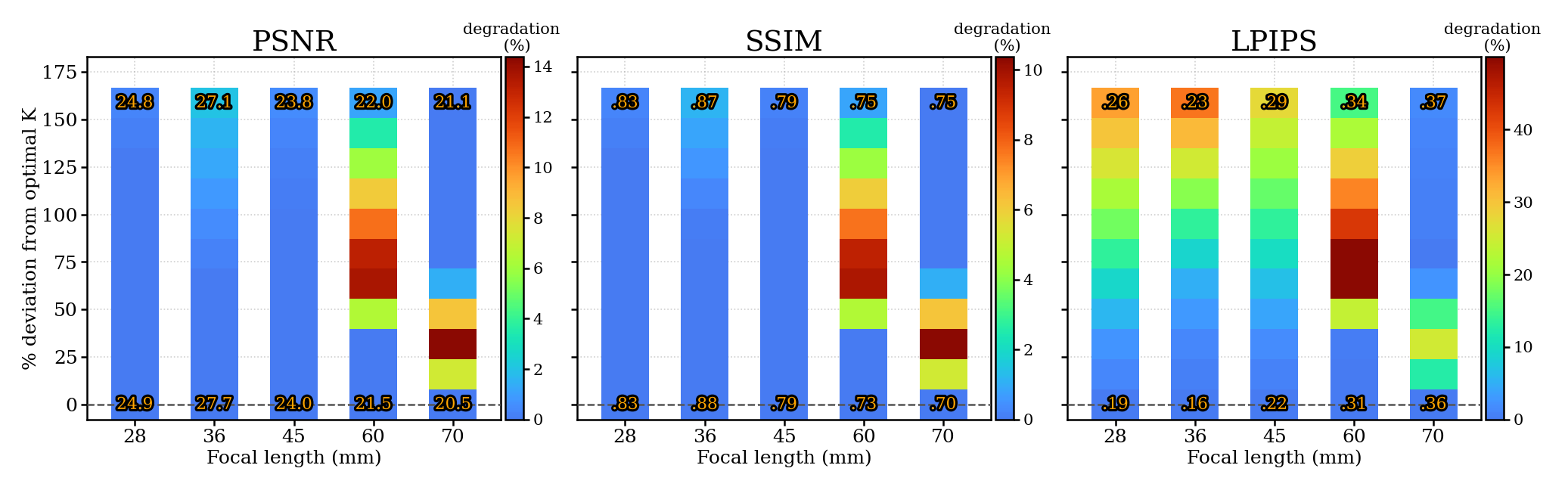}
     \vspace{-20pt}
    \caption{Bokehdiff sensitivity plot.}
\end{subfigure}\hfill

\caption{
Sensitivity Analysis for 4 DoF methods.}
\label{fig:sensitivity_all}
\end{figure}

We perform sensitivity analysis for all DoF methods to measure performance degradation when parameters deviate from their optimal values identified in Section \ref{subsec:tuning}. In Fig.~\ref{fig:sensitivity_all}, y-axis denotes \% deviation from the optimal parameter value. Color temperature denotes \% degradation in PSNR, SSIM or LPIPS metric. Printed values show optimal and degraded metric values. 

Bokehlicious, BokehMe and Dr.Bokeh show higher sensitivity to the parameter deviation at lower frequencies of 28-45mm. For Boekhlicious (Fig.~\ref{fig:sensitivity_all}(a)) at lower frequencies, the LPIPS degradation trend is super-linear: 50\% degradation when deviating 50\% from the optimal $\mathcal{F}$, then 200\% degradation when deviating 83\% from optimal $\mathcal{F}$. In contrast, Bokehlicious at higher frequencies and the other three methods at all frequencies show sub-linear degradation w.r.t. parameter deviation.  

Lower degradation at higher frequencies of 60-70mm can be attributed to heavy presence of narrow field of view (zoomed-in) images in the training datasets. All four models are trained on datasets with few objects in the narrow FoV per image at higher focal lengths. Bokehlicious paper explicitly mentions that the focal length is chosen per the photographer's descretion. Detailed look through training datasets for all four methods show high unbalance of focal lengths favoring narrow FoV performance. In sharp contrast, MODEST bears significant advantage with its balanced image distribution across 10 focal lengths.

This sensitivity analysis demonstrates how deviations from the optimal values of tunable parameters of each DoF model affect their performance. Clear trends with respect to focal length highlight performance-skew that favors select focal lengths and impacts end-user experience for others. With its balanced capture across 10 focal lengths, MODEST helps with such systemic evaluation and anomaly discovery.

% \section{Limitations}\label{sec:limitations}
% \textcolor{red}{
% add if a reviewer demands.
% }

\section{Conclusion}\label{sec:conclusion}
\vspace{-0.5ex}In this work, we introduced the first ultra-high-resolution (20MP), controlled-optics, balanced, real-world stereo dataset, MODEST, comprising 20,000 images across 10 scenes with challenging visual elements, designed for shallow depth of field (DoF) rendering and defocus deblurring. The proposed dataset is then systematically applied to assess models and benchmark the state-of-the-art depth of field and defocus deblurring methods. With quantitative and visual evidence, we demonstrate that existing DoF and deblurring models degrade substantially under real camera optics and struggle with elements such as reflective or semi-transparent surfaces and fine details. SOTA DoF methods fail to render realistic blur shapes and misidentify correct blur regions and blur amount. For defocus deblur methods, our quantitative evidence indicates that the methods have challenges with reconstructing sharp details for wide apertures and large focal lengths. We also conduct tuning studies and sensitivity analysis for the configurable parameters of the SOTA DoF methods. Three DoF methods, BokehMe, Dr.Bokeh, and BokehDiff, demonstrate insensitivity to the image focal attributes, while the Bokehlicious method has a nonlinear trend with the focal length at a constant aperture. This study shows the application of our ultra-high-resolution and optically structured dataset for assessment of SOTA methods and model improvements. To support future research, we release the complete dataset, illustrations, calibration images, data processing, and evaluation tools for non-commercial and academic research. With 20000 high-fidelity images and real systematic camera optics, we hope this dataset catalyzes advancing shallow depth of field and defocus deblurring towards more robust performance on real images and better model architectures with tunable parameters proportional to image focal attributes.

% ---------------------------------------------------------------
% Acknowledgments
% In JMLR, acknowledgments and funding go before the references and appendices.
% PUT THE ACKS BLOCK HERE
\section*{Acknowledgments and Disclosure of Funding}
{\vspace{-2ex}We self-funded the hardware, data acquisition setup, and cloud computing resources (e.g., Google Cloud, Google Colab) as required for this research work. This research is not a part of any specific grant from funding agencies in the public, commercial, or not-for-profit sectors. We declare no competing financial interests. We thank our peers and mentors for brainstorming, menuscript review and suggestions, and we thank human volunteers for their time and efforts ranking the depth of field methods for MoR evaluation in the benchmarking.}

%\clearpage    
% ---------------------------------------------------------------
% Bibliography
% JMLR sets the bibliographystyle to plainnat automatically. 
% You do NOT need \bibliographystyle{IEEEtran} here.
\bibliography{references}

\clearpage

% ---------------------------------------------------------------
% Appendices
\appendix
\section{Extra Experiments}

This section presents additional qualitative experiments that complement the main paper. We provide further visual comparisons across different tasks and evaluation settings to illustrate the behavior of the evaluated methods under a wider range of scenes and imaging conditions. These supplementary results reinforce the observations discussed in the main text and offer additional evidence of the strengths and limitations of the compared approaches.

\subsection{Shallow DoF Rendering Models}

\begin{figure}[!ht]  
    \centering
    \setlength{\abovecaptionskip}{1pt}  % reduce space above caption
    \setlength{\belowcaptionskip}{1pt}  % optional: reduce space below caption    
    \includegraphics[width=\textwidth]{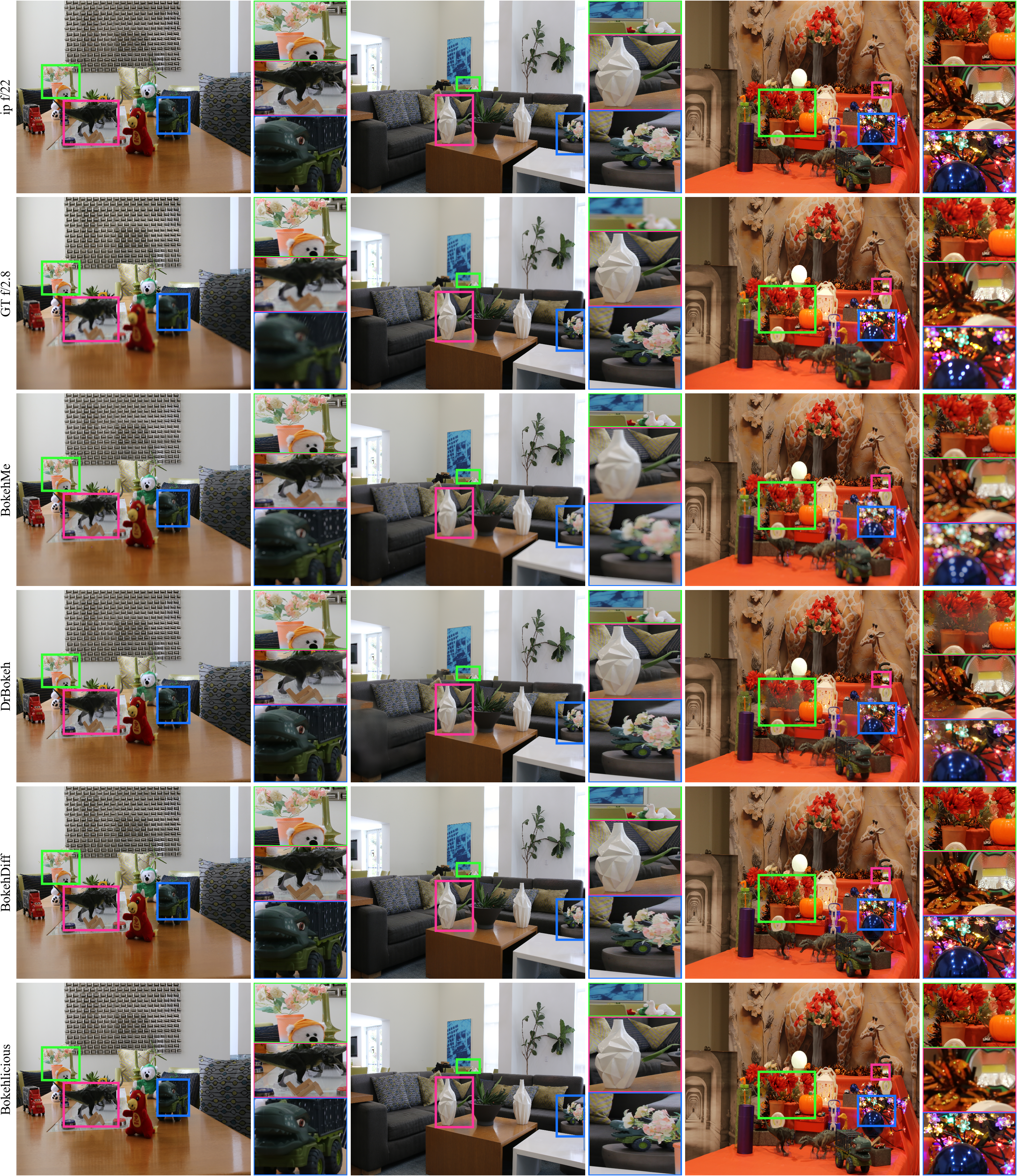}   
    % scale to text width    
    \caption{Additional examples of shallow depth of field (DoF) rendering methods for three different scenes for focal lengths of 55mm, 28mm, 70mm respectively. Sharp input is f/22.0  aperture, output target is f/2.8  aperture. Row 4, last column shows examples of gray color bleeding in Dr.Bokeh outputs, even though reducing color bleeding is one of the main contributions claimed in the Dr.Bokeh paper. Further, the last column also shows failure of all models in rendering cat-eye-shaped vignetting.}
    \label{fig:dof_suppl1}
\end{figure}

Fig.~\ref{fig:dof_suppl1} shows additional visual examples for shallow depth of field (DoF) rendering methods across variations of scenes and different focal lengths, including 55mm, 28mm, and 70mm. We use images captured by an f/22.0  aperture and an f/2.8  aperture for the model’s input and target outputs. The last four rows show shallow DoF images rendered by different DoF methods. The BokehMe method relatively successfully renders DoF outputs closest to GT compared with other shallow DoF rendering methods. In the second column, BokehMe successfully blurs the ceramic containers within pink bounding boxes but keeps the window areas sharp. The last column shows a failure case for all shallow depth of field (DoF) methods in rendering cat-eye shaped vignetting bokeh. Real camera images often have non-circular bokeh, or bokeh with varying color intensity. Current SOTA methods produce symmetric, circular, uniform density bokeh shapes which are far from realism.
%All the methods fail to blur light-emitting sources and a reflective sphere within blue bounding boxes. This example illustrates that MODEST contains complex scenes and increases the difficulty for shallow DoF tasks.

\subsection{Multi-Capture}

\begin{figure}[!ht]  
    \centering
    \setlength{\abovecaptionskip}{1pt}  % reduce space above caption
    \setlength{\belowcaptionskip}{1pt}  % optional: reduce space below caption    
    \includegraphics[width=0.7\textwidth]{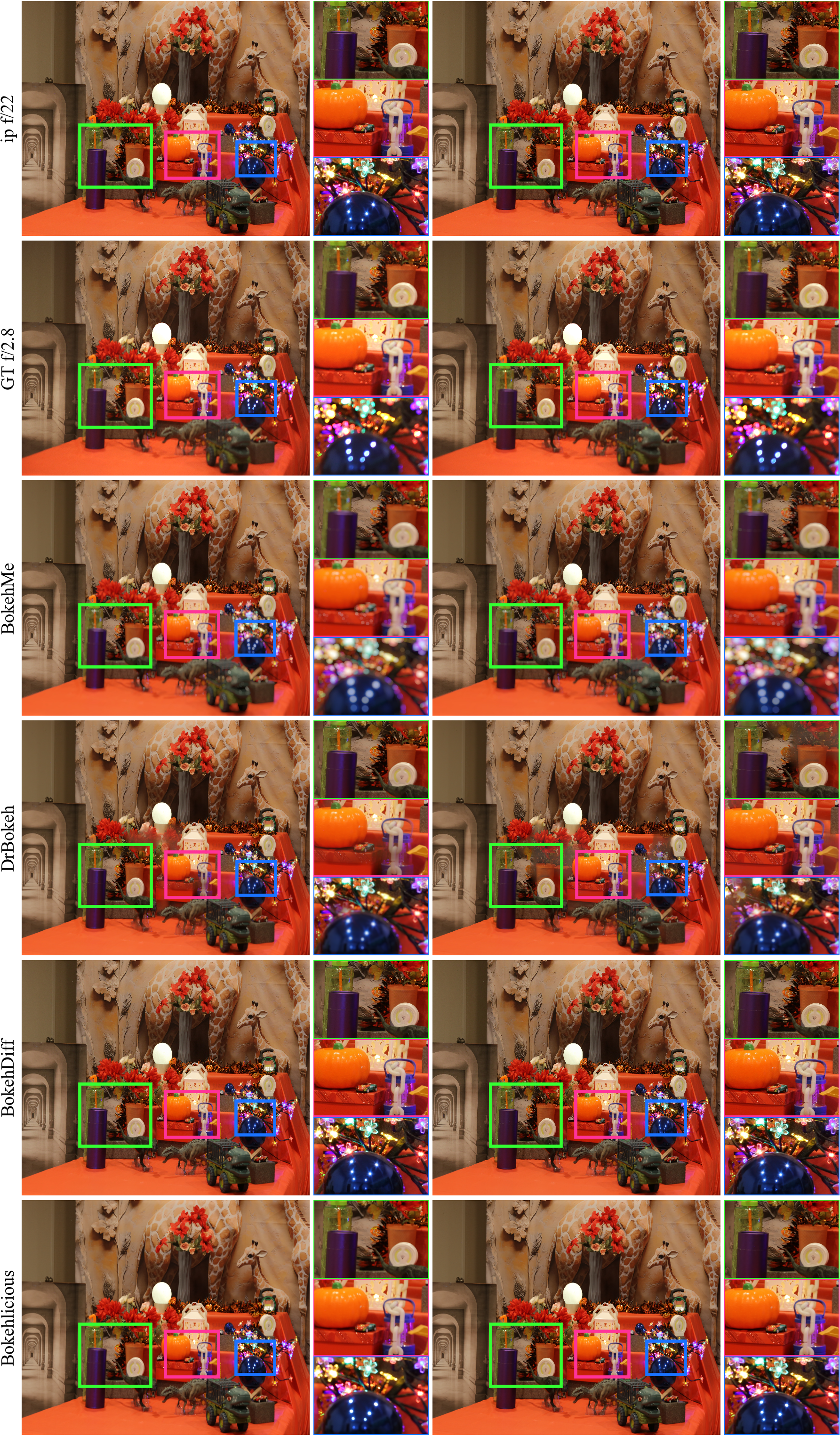}  
    % scale to text width    
    \caption{Performance of depth of field (DoF)methods on multi-capture images. MODEST dataset features at least 2 captures per viewpoint per focal configuration. That is, camera shutter is remotely operated twice consequtively, resulting into very minimal lens motion and near-identical two captures. These provide an excellent test-setting for depth of field methods under micro-perturbations. As demonstrated above, BokehMe appears more robust, while the rest of DoF methods produce significantly different outputs for near-identical inputs for the same model setting.}
    \label{fig:dof_suppl2_multicapture}
\end{figure}

Fig.~\ref{fig:dof_suppl2_multicapture} shows visual examples for shallow depth of field (DoF) rendering methods on multi-capture images. The MODEST dataset contains images captured by at least 2 shots per viewpoint per focal configuration. We use images captured by an f/22.0  aperture and an f/2.8  aperture for the model’s input and target outputs. The two columns show DoF outputs on multi-capture images. The DrBokeh and Bokehlicious F4.5 methods render significantly different DoF outputs on the multi-capture image. DrBokeh blurs light-emitting sources and a reflective sphere in one capture but renders them relatively sharp in the other capture. These examples demonstrate variations of model outputs caused by micro-perturbation for DoF methods. Compared to the other methods, BokehMe is relatively robust against micro-perturbation.

\subsection{Deblurring Models}

\begin{figure}[!ht]  
    \centering
    \setlength{\abovecaptionskip}{1pt}  % reduce space above caption
    \setlength{\belowcaptionskip}{1pt}  % optional: reduce space below caption    
    \includegraphics[width=\textwidth]{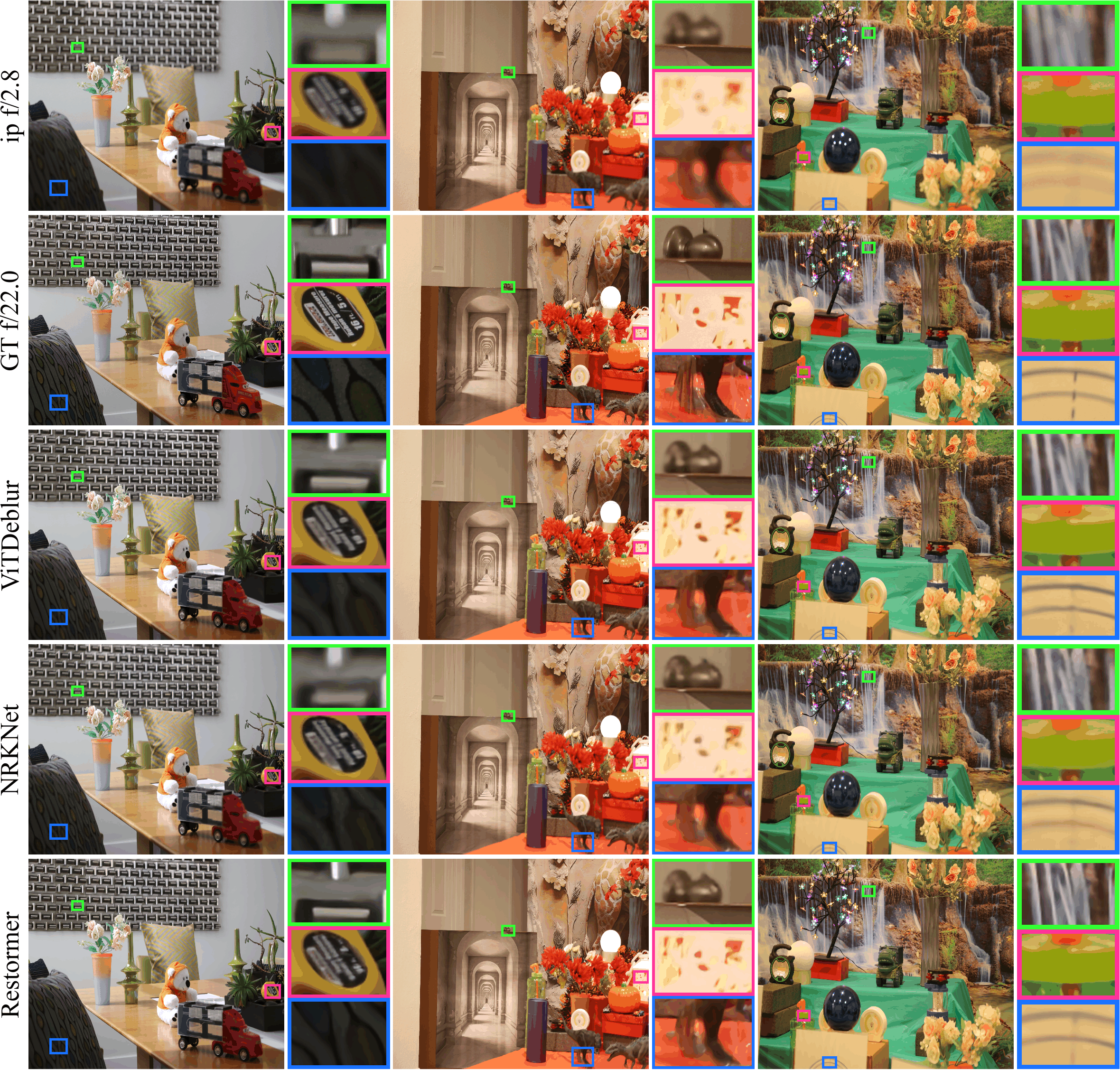}   
    % scale to text width    
    \caption{Additional qualitative comparison of defocus deblurring methods across three representative scenes captured at a focal length of 70 mm. The input image corresponds to an aperture of f/2.8, and the restored outputs are compared against the sharp reference captured at f/22.0. The three columns correspond to Scenes 5, 8, and 1, respectively. Zoomed-in regions highlight challenging failure cases, including degraded text readability, inaccurate restoration of thin structures, errors in object boundaries, patterns, and other fine geometric details }
    \label{fig:deblur_suppl1}        
\end{figure}

\begin{figure}[!ht] 
    \centering
    \setlength{\abovecaptionskip}{1pt}  % reduce space above caption
    \setlength{\belowcaptionskip}{1pt}  % optional: reduce space below caption    
    \includegraphics[width=\textwidth]{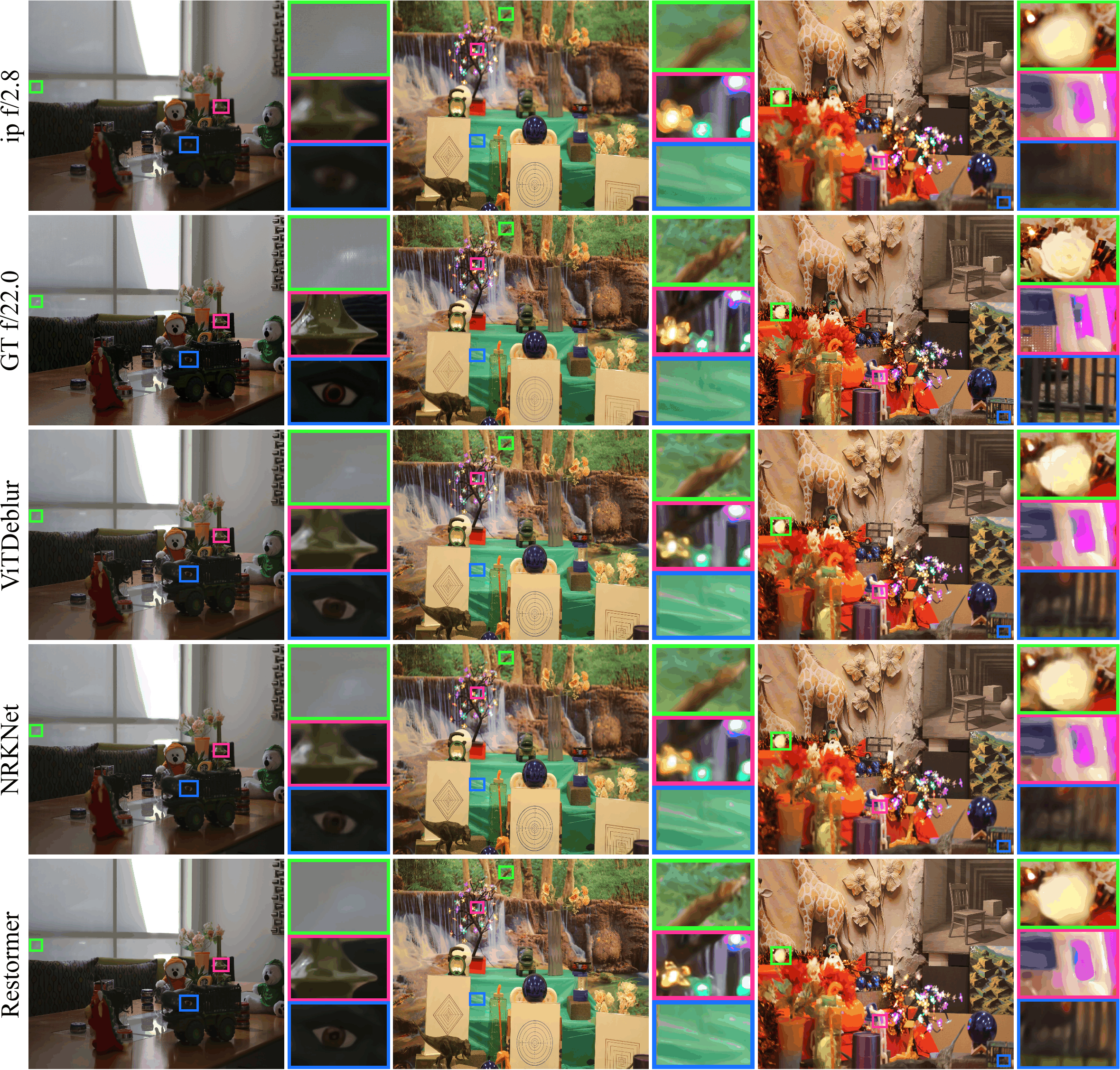}   
    % scale to text width    
    \caption{Additional qualitative examples of defocus deblurring results across three different scenes. Zoomed-in crops emphasize challenging regions where existing methods struggle to recover high-frequency details, including specular reflections, bright light sources, closely spaced objects (e.g., overlapping rose petals), complex textures, and fine structural details. These examples further illustrate the limitations of current deblurring approaches under diverse scene content and severe defocus blur.}
    \label{fig:deblur_suppl2}        
\end{figure}

Fig.~\ref{fig:deblur_suppl1} and Fig.~\ref{fig:deblur_suppl2} show additional visual examples for defocus deblurring methods across variations of scenes. We use images captured by an f/2.8  aperture and an f/22.0 aperture for the model’s input and target outputs. The first two rows represent inputs and outputs for the methods, and the remaining row reveals the deblurred outputs of the defocus deblurring methods. The ViTDeblur method relatively successfully sharpens defocused images from blurred regions to a sharp one across different scenes. Like ViTDeblur, the Restomer method also enables rendering sharp regions from blurred inputs. These examples demonstrate that our MODEST dataset poses a challenge for defocus deblurring methods due to the complex scenes with fine detail objects.

As shown in Fig.~\ref{fig:deblur_suppl1} and Fig.~\ref{fig:deblur_suppl2}, several failure cases are consistently observed across all methods. Models struggle to accurately handle fine structures such as flower petals, often failing to preserve spatially varying blur and resulting in either sharpening or loss of detail. Additionally, inconsistencies in deblurring light reflections across different surfaces indicate limitations.

\begin{figure}[!ht] 
    \centering
    \setlength{\abovecaptionskip}{1pt}
    \setlength{\belowcaptionskip}{1pt}    
    
    \includegraphics[width=\textwidth]{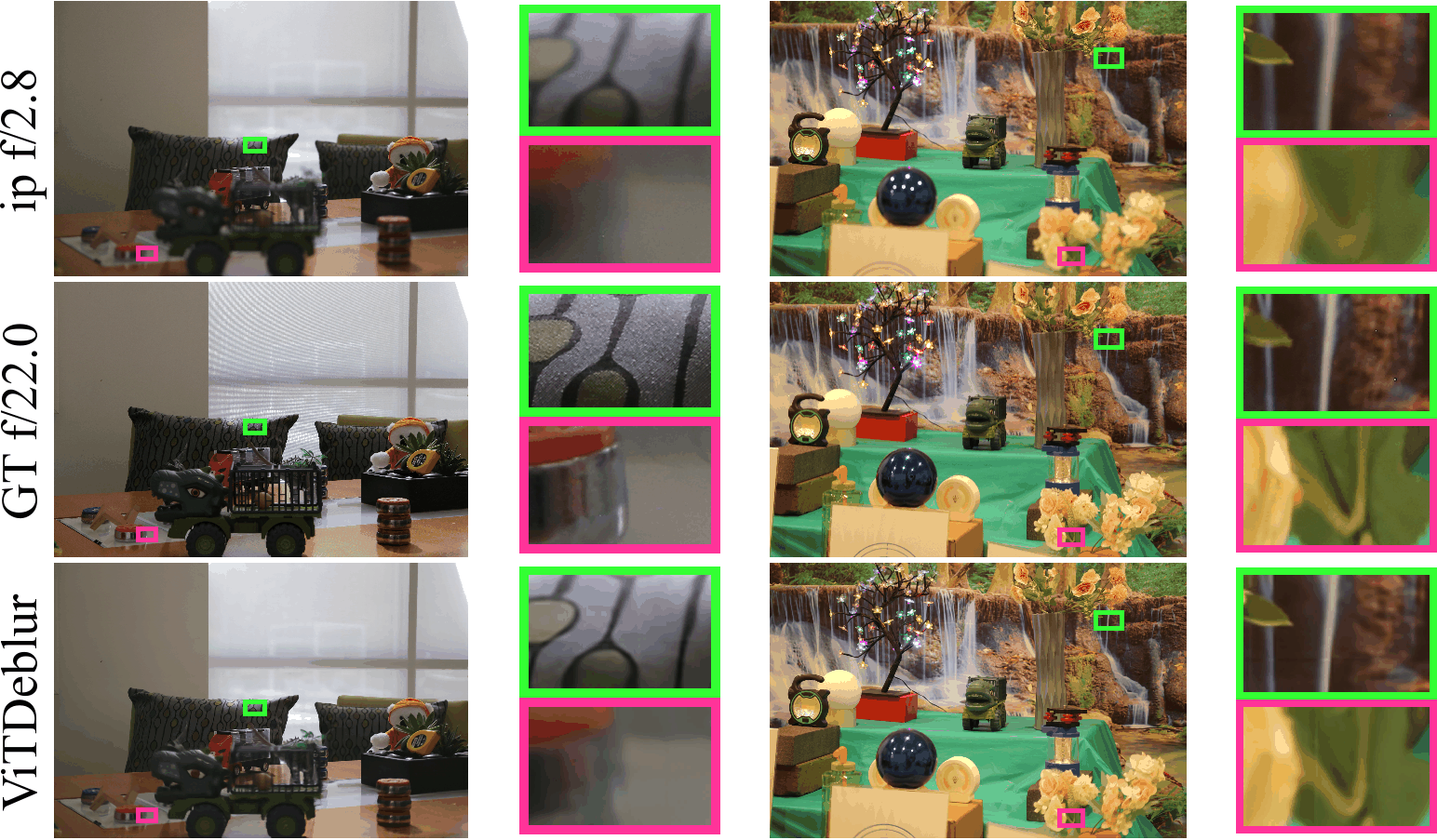}
    
    \caption{Zoomed-in visualization highlighting patch boundary artifacts in ViTDeblur outputs. Visible grid-like seams and local inconsistencies reveal incomplete blending between neighboring patches, resulting in non-uniform defocus restoration.}
    \label{fig:vit_deblur_only}        
\end{figure}

As illustrated in Fig.~\ref{fig:vit_deblur_only}, ViTDeblur exhibits prominent patching artifacts in zoomed regions. These artifacts appear as block-like discontinuities and local texture inconsistencies, suggesting limitations in smothening the patches during reconstruction. Such patching becomes particularly evident in regions with smooth gradients or subtle blur transitions, where the model fails to maintain continuity.NRKNet fails to correctly defocus certain regions, leading to unrealistic sharpness in areas that should remain blurred. Restormer, while generally more stable, shows degradation in highly zoomed-in regions, highlighting its limitations in preserving fine-grained details under extreme conditions.

\subsection{Ablation Study}

\begin{figure}[htbp] 
    \centering
    \setlength{\abovecaptionskip}{1pt}  % reduce space above caption
    \setlength{\belowcaptionskip}{1pt}  % optional: reduce space below caption    
    \includegraphics[width=\textwidth]{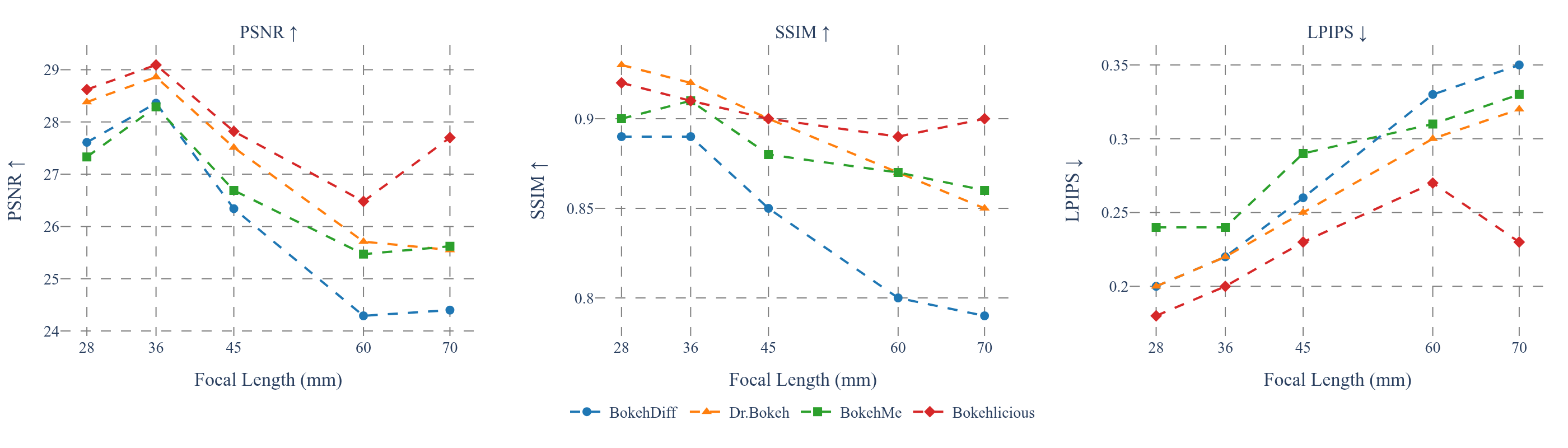}\\
    \includegraphics[width=\textwidth]{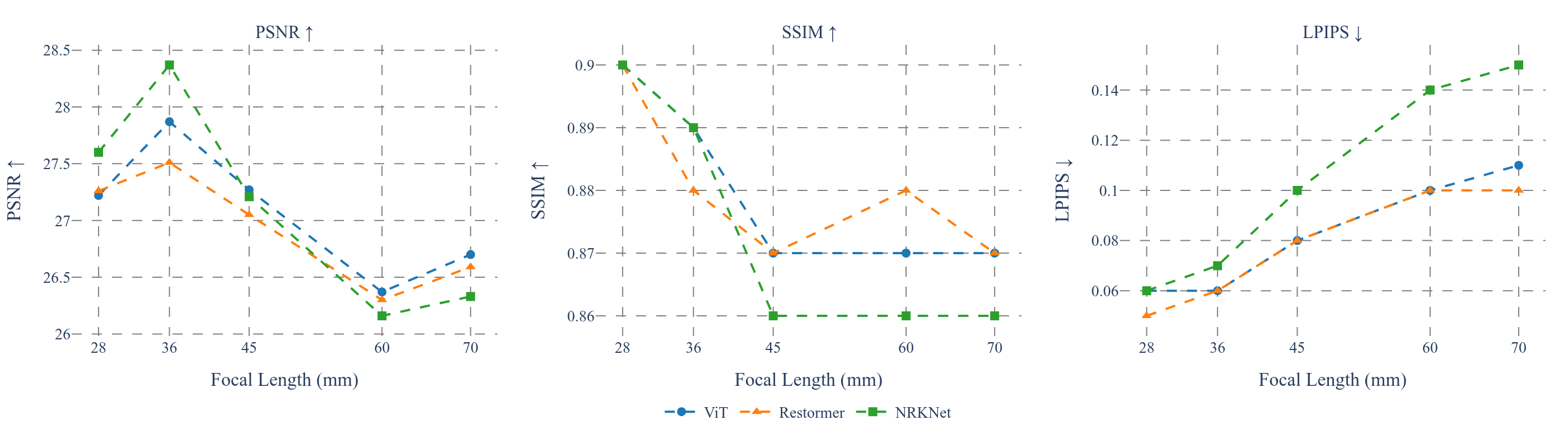}\\
    \includegraphics[width=\textwidth]{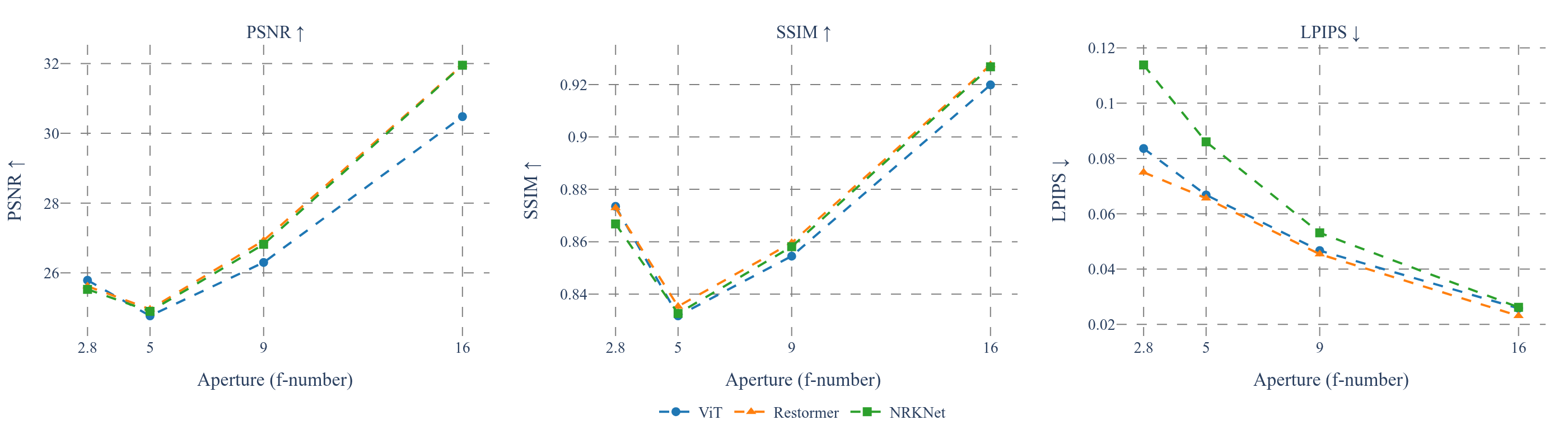}
    % scale to text width    
    \caption{Ablation study for 5 DoF and 3 defocus deblur methods. Top row shows depth of field methods vs focal length. Middle row shows defocus deblur methods vs focal length. Bottom row shows defocus deblur methods vs apertures.}
    \label{fig:deblur_dof_metrics}        
\end{figure}

The first row in Fig.~\ref{fig:deblur_dof_metrics} shows an ablation study across five different DoF rendering methods across variations of focal lengths (28mm to 70mm) using three standard image quality metrics: PSNR, SSIM, and LPIPS. BokehDiff maintains the highest PSNR and SSIM scores while keeping the lowest LPIPS scores at wide focal lengths (i.e., 28mm-36mm). It shows that this method produces the most accurate results. Both versions of the Bokehlicious methods consistently underperform compared to the rest. Compared to Bokehlicious-2.8, the 4.5 version significantly outperforms across all metrics.

The second row in Fig.~\ref{fig:deblur_dof_metrics} shows an ablation study across three different defocus deblurring methods across variations of focal lengths (28mm to 70mm) using three standard image quality metrics: PSNR, SSIM, and LPIPS. All three deblurring methods have peak performance at the 36mm focal length. There are the highest PSNR and SSIM scores and the lowest perceptual error (LPIPS). In addition, ViT and Restormer methods are much more stable, while NRKNet has worse results at wide angles. The LPIPS scores of these two methods are nearly identical, showing they produce a realistic look to the human eye at longer focal lengths.

The third row in Fig.~\ref{fig:deblur_dof_metrics} shows an ablation study across three different defocus deblurring methods across variations in apertures using three standard image quality metrics: PSNR, SSIM, and LPIPS. The Restormer method is the most reliable for a wide aperture (f/2.8 ). This method maintains the lowest LPIPS score. As the f-number increases, the performance gap among the three models almost diminishes. By f/16 the metrics for all methods are nearly identical. This indicates that method choice is less important when the input image is relatively sharp.

In figure ~\ref{fig:deblur_dof_metrics},  an interesting observation is found across deblurring results. Although f/2.8  corresponds to stronger blur than f/5.0 , the performance drop in terms of PSNR and SSIM is more pronounced at f/5.0, Intuitively, heavier blur at  should result in lower reconstruction quality; however, the results suggest a non-monotonic relationship between blur strength and restoration performance. This indicates that model behavior is not solely determined by blur magnitude. Instead, intermediate blur levels such as f/5.0  may introduce more complex or less structured degradations that are harder for models to invert. Additionally, factors such as scene-dependent blur patterns, texture distribution, and model bias toward extreme cases may further contribute to this unexpected trend.

\subsection{Calibration Results}

Tables~\ref{tab:monocal_eos6d_a} and~\ref{tab:monocal_eos6d_b} summarize the intrinsic calibration results for Camera-1 and Camera-2, respectively. Intrinsic parameters were estimated independently for each focal length using the provided ChArUco calibration images. For every focal length, the estimated focal lengths, principal point, reprojection RMS error, and radial and tangential distortion coefficients are reported. The consistently low reprojection errors across the full zoom range indicate accurate and stable intrinsic calibration.

Monocular calibration was performed using OpenCV's ChArUco calibration pipeline with a custom $12\times16$ ChArUco board consisting of $45\,\mathrm{mm}$ squares and DICT\_4X4\_100 markers. 50 images were captured per focal length from diverse viewing angles and distances to maximize geometric coverage. ChArUco corners were detected and camera intrinsics were estimated using the standard pinhole camera model while jointly estimating the radial distortion coefficients ($k_1$, $k_2$) and tangential distortion coefficients ($p_1$, $p_2$). The third radial distortion coefficient ($k_3$) was fixed to zero during optimization for improved numerical stability.

Stereo calibration was subsequently performed independently for each focal length using stereo checkerboard captures, while keeping the previously calibrated intrinsics fixed for each camera. Following an initial stereo calibration, images exhibiting inconsistent baseline estimates or high epipolar alignment error were automatically rejected before a second refinement step. The final stereo parameters were then used to compute rectification transforms and dense remapping functions, ensuring accurate epipolar alignment for all released stereo image pairs.
% \section{Calibration}\label{sec:calibration_detail}

\begin{table}[htbp]
\centering
\caption{Estimated intrinsic camera parameters obtained from monocular ChArUco calibration of Camera-1. Results include focal lengths ($f_x$, $f_y$), principal point ($c_x$, $c_y$), reprojection RMS error, and radial ($k_1$, $k_2$) and tangential ($p_1$, $p_2$) distortion coefficients for each focal length.}
\label{tab:monocal_eos6d_a}
\resizebox{0.8\textwidth}{!}{%
\begin{tabular}{cccccccccccc}
\hline 
\textbf{FL} & \textbf{$f_x$} & \textbf{$f_y$} & \textbf{$f_y$/$f_x$} & \textbf{$c_x$} & \textbf{$c_y$} & \textbf{Reprojection} & \textbf{$k_1$} & \textbf{$k_2$} & \textbf{$p_1$} & \textbf{$p_2$} \\
\textbf{(mm)} & \textbf{(px)} & \textbf{(px)} & & \textbf{(px)} & \textbf{(px)} & \textbf{RMS(px)} & & & & \\
\hline
28  & 4450.87  & 4466.40  & 1.0035 & 2712.2 & 1804.3 & 0.0742 & $-$0.10635 & $+$0.12080 & $-$0.000828 & $-$0.001525   \\
32  & 4852.57  & 4870.70  & 1.0037 & 2724.1 & 1803.8 & 0.0690 & $-$0.10605 & $+$0.14133 & $-$0.000749 & $-$0.000706   \\
36  & 5516.57  & 5537.89  & 1.0039 & 2715.7 & 1802.5 & 0.0766 & $-$0.09400 & $+$0.15951 & $-$0.001084 & $-$0.001363   \\
40  & 6144.46  & 6167.70  & 1.0038 & 2714.5 & 1806.2 & 0.0786 & $-$0.07970 & $+$0.18767 & $-$0.001113 & $-$0.001703   \\
45  & 6925.18  & 6950.42  & 1.0036 & 2709.3 & 1798.9 & 0.0749 & $-$0.05054 & $+$0.18603 & $-$0.001492 & $-$0.001956   \\
50  & 7842.56  & 7871.89  & 1.0037 & 2690.3 & 1798.3 & 0.0876 & $-$0.01658 & $+$0.21962 & $-$0.001805 & $-$0.002670   \\
55  & 8572.39  & 8602.62  & 1.0035 & 2680.1 & 1781.5 & 0.0937 & $+$0.02604 & $+$0.17361 & $-$0.002671 & $-$0.003194   \\
60  & 9273.11  & 9308.46  & 1.0038 & 2685.2 & 1791.1 & 0.0947 & $+$0.06423 & $+$0.14536 & $-$0.002394 & $-$0.002865   \\
65  & 10075.31 & 10111.06 & 1.0035 & 2592.5 & 1765.0 & 0.0748 & $+$0.11405 & $+$0.03917 & $-$0.003637 & $-$0.007653   \\
70  & 10441.99 & 10476.16 & 1.0033 & 2652.9 & 1759.9 & 0.0929 & $+$0.13246 & $+$0.00550 & $-$0.003539 & $-$0.004344   \\
\hline
\end{tabular}%
}
\end{table}

\begin{table}[htbp]
\centering
\caption{Estimated intrinsic camera parameters obtained from monocular ChArUco calibration of Camera-2. Results include focal lengths ($f_x$, $f_y$), principal point ($c_x$, $c_y$), reprojection RMS error, and radial ($k_1$, $k_2$) and tangential ($p_1$, $p_2$) distortion coefficients for each focal length.}
\label{tab:monocal_eos6d_b}
\resizebox{0.8\textwidth}{!}{%
\begin{tabular}{ccccccccccccc}
\hline
\textbf{FL} & \textbf{$f_x$} & \textbf{$f_y$} & \textbf{$f_y$/$f_x$} & \textbf{$c_x$} & \textbf{$c_y$} & \textbf{Reprojection} & \textbf{$k_1$} & \textbf{$k_2$} & \textbf{$p_1$} & \textbf{$p_2$} \\
\textbf{(mm)} & \textbf{(px)} & \textbf{(px)} & & \textbf{(px)} & \textbf{(px)} & \textbf{RMS(px)} & & & &\\
\hline
28  & 4452.00  & 4467.49  & 1.0035 & 2729.7 & 1815.5 & 0.0653          & $-$0.10821 & $+$0.11935 & $-$0.000482 & $-$0.000170 \\
32  & 4824.70  & 4841.87  & 1.0036 & 2737.7 & 1816.1 & 0.0728          & $-$0.10511 & $+$0.13454 & $-$0.000238 & $+$0.000045 \\
36  & 5572.01  & 5592.07  & 1.0036 & 2715.9 & 1801.2 & 0.0771          & $-$0.09308 & $+$0.16312 & $-$0.001123 & $-$0.001591 \\
40  & 6076.63  & 6094.92  & 1.0030 & 2719.6 & 1788.0 & 0.0911          & $-$0.07649 & $+$0.16438 & $-$0.001430 & $-$0.001300 \\
45  & 6852.10  & 6877.25  & 1.0037 & 2686.7 & 1793.0 & 0.0816          & $-$0.05010 & $+$0.17375 & $-$0.001542 & $-$0.003267 \\
50  & 7704.99  & 7732.78  & 1.0036 & 2705.8 & 1785.9 & 0.0992          & $-$0.02153 & $+$0.21945 & $-$0.001836 & $-$0.002140 \\
55  & 8431.15  & 8457.85  & 1.0032 & 2751.3 & 1807.2 & 0.0596          & $+$0.01268 & $+$0.22296 & $-$0.000900 & $-$0.000682  \\
60  & 9156.86  & 9189.64  & 1.0036 & 2708.6 & 1765.4 & 0.0932          & $+$0.04977 & $+$0.17149 & $-$0.002751 & $-$0.002504 \\
65  & 9947.65  & 9984.50  & 1.0037 & 2636.5 & 1765.8 & 0.0912          & $+$0.09036 & $+$0.21429 & $-$0.002900 & $-$0.005800 \\
70  & 10470.27 & 10502.91 & 1.0031 & 2664.8 & 1727.8 & 0.0684          & $+$0.11583 & $+$0.20310 & $-$0.003475 & $-$0.004875 \\
\hline
\end{tabular}%
}
\end{table}

\end{document}